\def\year{2017}\relax
\documentclass[letterpaper]{article}
\usepackage{aaai17}
\usepackage{times}
\usepackage{helvet}
\usepackage{courier}

\usepackage{subcaption}
\usepackage{graphicx}
\usepackage{mathtools}
\usepackage{color}
\usepackage[ruled]{algorithm2e}
\usepackage{amssymb,amsmath}
\usepackage{amsthm}

\def\given{\;\middle\vert\;}

\def\indicator{\mathbf{1}}
\DeclareMathOperator*{\expectation}{\mathbb{E}}
\DeclareMathOperator{\prob}{P}
\def\actions{\mathcal{A}}
\def\options{\Omega}
\def\states{\mathcal{S}}

\def\defeq{=}\newcommand{\deriv}[2][]{\frac{\partial#1}{\partial#2}}
\newtheorem{theorem}{Theorem}

\def\option{\omega}

\def\termparams{\vartheta}

\DeclareMathOperator{\argmax}{argmax}

\begin{document}
\title{The Option-Critic Architecture}
\author{Pierre-Luc Bacon \and Jean Harb \and Doina Precup \\
Reasoning and Learning Lab, School of Computer Science \\
McGill University\\
\texttt{\{pbacon, jharb, dprecup\}@cs.mcgill.ca}\\
}
\nocopyright
\maketitle
\begin{abstract}
Temporal abstraction is key to scaling up
learning and planning in reinforcement learning. While planning
with temporally extended actions  is well understood,
creating such abstractions autonomously from data has remained challenging. We tackle this problem
in the framework of options [Sutton, Precup \& Singh, 1999; Precup, 2000].
We derive policy gradient theorems for options and propose a new
{\em option-critic} architecture capable of learning both the
internal policies and the termination conditions of options, in tandem
with the policy over options, and without the need to provide any additional
rewards or subgoals. Experimental results
in both discrete and continuous environments showcase the flexibility and
efficiency of the framework.
 \end{abstract}

\section{Introduction}
Temporal abstraction allows representing
knowledge about courses of action that take place  at different time scales.
In reinforcement learning,
\textit{options}~\cite{Sutton1999opt,precup00} provide a framework for defining
such courses of action and for seamlessly learning and planning with them.
{\em Discovering} temporal abstractions autonomously
has been the subject of extensive research efforts in the last 15 years
~\cite{McGovern2001,Stolle2002,Menache2002,Simsek08,Silver2012},
but approaches that can be used naturally with continuous state and/or action spaces have only recently
started to become feasible~\cite{Konidaris2011,Niekum2013,Mann2015,Mankowitz2016,Kulkarni2016,Vezhnevets2016,Daniel2016}.

The majority of the existing work has focused on
finding {\em subgoals} (useful states that an agent should reach) and subsequently
learning policies to achieve them.
This idea has led to interesting methods but ones which are also difficult
to scale up given their  ``combinatorial'' flavor.
Additionally, learning policies associated with  subgoals can be expensive
in terms of data and computation time; in the worst case, it can be as expensive
as solving the entire task.

We present an alternative view, which blurs the line between
the problem of {\em discovering options} from that of {\em learning options}.
Based on the policy gradient theorem \cite{Sutton1999pg}, we derive new results
which enable a gradual learning process of the intra-option policies and
termination functions, simultaneously with the policy over them. This approach
works naturally with both linear and non-linear function
approximators, under discrete or continuous state and action spaces. Existing
methods for learning options are considerably slower when learning from a
single task: much of the benefit comes from re-using
the learned options in similar tasks. In contrast, we show that our approach
is capable of successfully learning options within a single task
without incurring any slowdown and while still providing benefits for transfer learning.

We start by reviewing background related to the two main ingredients of our work: policy gradient methods and options.
We then describe the core ideas of our approach: the intra-option policy
and termination gradient theorems. Additional technical details are included in
the appendix. We present experimental results
showing that our approach learns meaningful temporally extended
behaviors in an effective manner. As opposed
to other methods, we only need to specify the number of desired
options; it is not necessary to have subgoals, extra rewards, demonstrations,
multiple problems or any other special accommodations (however, the approach can
take advantage of pseudo-reward functions if desired).
To our knowledge, this is the first end-to-end approach for learning options
that scales to very large domains at comparable efficiency.
 \section{Preliminaries and Notation} \label{sect:prelim}

A Markov Decision Process consists of a set of states $\mathcal{S}$,
a set of actions $\mathcal{A}$, a transition function
$\prob : \mathcal{S} \times \mathcal{A} \to (\mathcal{S} \to [0,1])$
and a reward function $r: \mathcal{S} \times \mathcal{A} \to \mathbb{R}$.
For convenience, we develop our ideas assuming discrete state and action sets.
However, our results extend to continuous spaces using usual measure-theoretic assumptions
(some of our empirical results are in continuous tasks).
A (Markovian stationary) \textit{policy}  is a probability
distribution over actions conditioned on states, $\pi: \mathcal{S} \times \mathcal{A} \to [0,1]$.
In discounted problems, the
value function of a policy $\pi$ is defined as the expected return:
$V_\pi(s) = \expectation_{\pi}\left[ \sum_{t=0}^\infty \gamma^{t} r_{t+1} \given s_0=s \right]$
and its action-value function as $Q_\pi(s, a) = \expectation_{\pi}\left[ \sum_{t=0}^\infty \gamma^{t} r_{t+1} \given s_0=s, a_0 = a \right]$,
where $\gamma \in [0, 1)$ is the \textit{discount factor}. A policy $\pi$ is
 \textit{greedy} with respect to a given action-value function $Q$ if $\pi(s,a)>0 \mbox{ iff } a=\argmax\limits_{a'} Q(s, a')$.
In a discrete MDP, there is at least one optimal policy which is greedy with
respect to its own action-value function.

{\bf Policy  gradient methods}~\cite{Sutton1999pg,Konda1999}
address the problem of finding a good policy by performing stochastic gradient
descent to optimize a performance objective over a given family of
parametrized stochastic policies, $\pi_{\theta}$. The policy gradient theorem~\cite{Sutton1999pg}
provides expressions for the gradient of the average reward and discounted reward
objectives with respect to $\theta$. In the discounted setting, the objective is
defined with respect to a designated start state (or distribution) $s_0$:
$\rho(\theta, s_0) = \expectation_{\pi_{\theta}}\left[ \sum_{t=0}^\infty \gamma^{t} r_{t+1} \given s_0 \right]$.
The policy gradient theorem shows that: $\deriv[\rho(\theta, s_0)]{\theta} = \sum_{s} \mu_{\pi_\theta}\left( s \given s_0 \right) \sum_a \deriv[ \pi_\theta \left(a | s \right)]{\theta} Q_{\pi_\theta}(s, a)$,
where $\mu_{\pi_\theta}\left(s \given s_0\right) = \sum_{t=0}^\infty \gamma^t \prob\left( s_t = s \given s_0 \right)$ is
a discounted weighting of the states along the trajectories starting from $s_0$.
In practice, the policy gradient is estimated from samples along the on-policy stationary
distribution. \cite{thomas2014bias} showed that neglecting the discount factor
in this stationary distribution makes the usual policy gradient estimator biased.
However, correcting for this discrepancy also reduces data efficiency.
For simplicity, we build on the framework of \cite{Sutton1999pg} and
discuss how to extend our results according to \cite{thomas2014bias}.

{\bf The options framework}~\cite{Sutton1999opt,precup00} formalizes the idea of
temporally extended actions.
A Markovian option $\omega \in \Omega$ is a triple $( \mathcal{I}_\omega, \pi_\omega, \beta_\omega )$
in which $\mathcal{I}_\omega \subseteq \mathcal{S}$ is an initiation set,
$\pi_\omega$ is an \textit{intra-option} policy, and $\beta_\omega: \mathcal{S} \to [0,1]$
is a termination function.
We also assume that $\forall s \in \states, \forall \omega \in \Omega: s \in \mathcal{I}_\omega $
(i.e., all options are available everywhere), an assumption made in the majority
of option discovery algorithms. We will discuss how to dispense with this
assumption in the final section.
\cite{Sutton1999opt,precup00} show that an MDP endowed with a set of options
becomes a Semi-Markov Decision Process~\cite[chapter 11]{Puterman1994},
which has a corresponding optimal value function over options $V_{\Omega}(s)$
and option-value function $Q_{\Omega}(s, \option)$.
Learning and planning algorithms for MDPs have their counterparts in this setting.
However, the existence of the underlying MDP offers the possibility of learning
about many different options in parallel : this is the idea of \textit{intra-option learning},
which we leverage in our work.
 \section{Learning Options}\label{sect:learning}
We adopt a  continual perspective on the problem of learning
options. At any time, we would like to distill all of the available
experience into every component of our system: value function and policy over
options, intra-option policies and termination functions.
To achieve this goal, we focus on learning option policies and termination
functions, assuming they are represented using differentiable parameterized
function approximators.

We consider the \textit{call-and-return} option execution model,
in which an agent picks option $\omega$ according to its policy over options
$\pi_\Omega$ , then follows the intra-option policy $\pi_{\omega}$ until termination
(as dictated by $\beta_{\omega}$), at which point this procedure is repeated.
Let $\pi_{\option, \theta}$ denote the intra-option policy of option
$\option$ parametrized by $\theta$ and  $\beta_{\option, \vartheta}$,
the termination function of $\option$ parameterized by $\vartheta$.
We present two new results for learning options,
obtained using as blueprint the policy gradient theorem~\cite{Sutton1999pg}.
Both results are derived under the assumption that the goal is to learn options
that maximize the expected return in the current task. However, if one wanted to add
extra information to the objective function, this could readily be done so long
as it comes in the form of an additive differentiable function.

Suppose we aim to optimize directly the discounted return, expected over all
the trajectories starting at a designated state $s_0$ and option $\omega_0$, then:
$\rho(\Omega, \theta, \vartheta, s_0, \omega_0) \defeq \expectation_{\Omega, \theta, \omega}\left[\sum_{t=0}^\infty\gamma^{t} r_{t+1} \given s_0, \omega_0\right]$.
Note that this  return depends on the policy over options, as well as the
parameters of the option policies and termination functions.
We will take gradients of this objective with respect to $\theta$ and $\vartheta$.
In order to do this, we will manipulate equations similar to those used in
\textit{intra-option} learning \cite[section 8]{Sutton1999opt}.
Specifically, the definition of the option-value function can be written as:
\begin{align}
Q_\options(s, \option)  =\sum_a \pi_{\option, \theta}\left( a \given s \right)Q_U(s, \option, a)\enspace , \label{eq:q}
\end{align}
where $Q_U : \states \times \options \times \actions \to \mathbb{R}$ is the value of executing an action in the context of a state-option pair:
\begin{align}
Q_U(s, \omega, a) &\defeq r(s,a) + \gamma \sum_{s'} \prob\left( s' \given s, a \right)U(\omega, s') \enspace . \label{eq:qswa}
\end{align}
Note that the $(s,\omega)$ pairs lead to an
augmented state space, cf.~\cite{levy2011}.  However, we will not work
explicitly with this space; it is used only to simplify the derivation.
The function $U:\Omega \times \states \rightarrow \mathbb{Re}$ is called the
option-value function \textit{upon arrival}, \cite[equation 20]{Sutton1999opt}.
The value of executing $\omega$ upon entering a state $s'$ is given by:
\begin{align}
\hspace{-1em}U(\omega, s')&\defeq (1 - \beta_{\omega, \vartheta}(s')) Q_\Omega(s', \omega) + \
\beta_{\omega, \vartheta}(s') V_\Omega(s') \hspace{-8pt}\label{eq:u}
\end{align}
Note that $Q_U$ and $U$ both depend on $\theta$ and $\vartheta$, but we do not
include these in the notation for clarity.
The last ingredient required to derive policy gradients is the Markov chain along which
the performance measure is estimated. The natural approach is to consider the
chain defined in the augmented state space, because state-option pairs now play
the role of regular states in a usual Markov chain. If option $\omega_t$ has been
initiated or is executing at time $t$ in state $s_t$,
then the probability of transitioning to $(s_{t+1}, \omega_{t+1})$ in
one step is:
\begin{align}
&\prob\left(s_{t+1}, \omega_{t+1} \given s_t, \omega_t \right) = \sum_{a} \pi_{\omega_t,\theta}\left(a \given s_t \right) \prob(s_{t+1} |\, s_t, a)( \notag\\
&(1 - \beta_{\omega_t,\vartheta}(s_{t+1}))\indicator_{\omega_{t}=\omega_{t+1}} + \
\beta_{\omega_{t},\vartheta}(s_{t+1})\pi_\Omega(\omega_{t+1} |\, s_{t+1})) \label{eq:chainstructure}
\end{align}

Clearly, the process given by \eqref{eq:chainstructure} is homogeneous. Under mild conditions,
and with options available everywhere, it is in fact ergodic, and a unique stationary
distribution over state-option pairs exists.

We will now compute the gradient of the expected discounted return with respect to
the parameters $\theta$ of the intra-option policies, assuming that they are stochastic
and differentiable.
From (\ref{eq:q}~, \ref{eq:qswa}), it follows that:
\begin{align*}
\deriv[Q_\Omega(s, \omega)]{\theta} &= \
\left(\sum_a \deriv[\pi_{\omega,\theta}\left( a \given s \right)]{\theta}  Q_U(s,\omega, a) \right) \\
&+ \sum_a \pi_{\omega,\theta}\left(a \given s\right)  \sum_{s^\prime} \gamma \prob\left( s^\prime \given s, a\right) \deriv[U(\omega, s')]{\theta}  .
\end{align*}
We can further expand the right hand side using
\eqref{eq:u} and \eqref{eq:chainstructure}, which yields the following theorem:

\begin{theorem}[Intra-Option Policy Gradient Theorem]\label{thm:intrapg}

Given a set of Markov options with stochastic intra-option policies differentiable
in their parameters $\theta$, the gradient of the expected discounted return
with respect to $\theta$ and initial condition $(s_0,\omega_0)$ is:
\begin{align*}
&\sum_{s, \omega} \mu_\Omega\left(s, \omega  \given s_0, \omega_0\right) \sum_a \deriv[\pi_{\omega,\theta}\left( a \given s \right)]{\theta} Q_U(s,\omega, a)\enspace ,
\end{align*}
where $\mu_\Omega\left(s, \omega \given s_0, \omega_0\right)$ is a discounted
weighting of state-option pairs along  trajectories starting from $(s_0, \omega_0)$:
$\mu_\Omega\left(s, \omega \given s_0, \omega_0\right) \defeq \sum_{t=0}^\infty \gamma^t \prob\left( s_t = s, \omega_t = \omega \given s_0, \omega_0 \right)$.
\end{theorem}
The proof is in the appendix.
This gradient describes the effect of a local change at the
primitive level on the global expected discounted return. In contrast, subgoal
or pseudo-reward methods assume the objective of an option is simply to optimize its
own reward function, ignoring how a proposed change would propagate
in the overall objective.

We now turn our attention to  computing gradients for the
termination functions, assumed this time to be stochastic and differentiable in
$\termparams$. From (\ref{eq:q},~\ref{eq:qswa},~\ref{eq:u}), we have:
\begin{align*}
\deriv[Q_\Omega(s, \omega)]{\vartheta} &= \sum_a \pi_{\omega,\theta}\left(a \given s\right)  \sum_{s^\prime} \gamma \prob\left( s^\prime \given s, a\right) \deriv[U(\omega, s')]{\vartheta} .
\end{align*}
Hence, the key quantity is the gradient of $U$. This is a natural consequence of the
call-and-return execution,  in which the ``goodness'' of
termination functions can only be evaluated  upon entering the next state.
The relevant gradient can be further expanded as:

\begin{align}
\deriv[U(\option, s')]{\termparams} &=-\deriv[\beta_{\option, \termparams}(s')]{\termparams} A_\options(s', \option) \, + \notag\\
&\gamma \sum_{\option'} \sum_{s''} \prob\left(s'', \option' \given s', \omega\right) \deriv[U(\option', s'')]{\termparams} \enspace, \label{eq:du-recursive}
\end{align}
where $A_{\Omega}$  is the advantage function~\cite{Baird1993} over options
$A_\Omega(s', \omega) = Q_\Omega(s', \omega) - V_\Omega(s')$. Expanding $ \deriv[U(\option', s'')]{\termparams} $ recursively leads to a
similar form as in theorem \eqref{thm:intrapg} but where the weighting of state-option
pairs is now according to a Markov chain shifted by
one time step: $\mu_\Omega\left( s_{t+1}, \omega_t \given s_t, \omega_{t-1}\right)$ (details are in the appendix).
\begin{theorem}[Termination Gradient Theorem]\label{thm:term}
Given a set of Markov options with stochastic termination functions differentiable
in their parameters $\vartheta$, the gradient of the expected discounted return objective
with respect to $\vartheta$ and the initial condition $(s_1, \omega_0)$ is:
\begin{align*}
-\sum_{s', \omega} \mu_\Omega\left(s', \omega \given s_1, \omega_0\right) \deriv[\beta_{\option,\termparams}(s')]{\termparams} A_\options(s', \option) \enspace ,
\end{align*}
where $\mu_\options\left(s', \omega \given s_1, \omega_0\right)$ is a discounted weighting of state-option pairs
from $(s_1, \omega_0)$: $\mu_\Omega\left(s, \omega \given s_1, \omega_0\right) \defeq \sum_{t=0}^\infty \gamma^t \prob\left( s_{t+1} = s, \omega_t = \omega \given s_1, \omega_0 \right)$.
\end{theorem}

The advantage function often appears in policy gradient methods
\cite{Sutton1999pg} when forming a \textit{baseline} to
reduce the variance in the gradient estimates. Its presence in that context has to do
mostly with  algorithm design.  It is interesting that in our case,
it follows as a direct consequence of the derivation and gives the theorem
an intuitive interpretation: when the option choice is suboptimal with respect
to the expected value over all options, the  advantage function is negative and it drives
the gradient corrections up, which increases  the odds of terminating. After termination,
the agent has the opportunity to pick a better option using $\pi_\Omega$.
A similar idea also underlies the \textit{interrupting} execution model of
options~\cite{Sutton1999opt} in which termination is forced whenever the value
of $Q_\Omega(s', \omega)$ for the current option $\omega$ is less than $V_\Omega(s')$.
\cite{Mann2014} recently studied interrupting options through the lens of an
\textit{interrupting Bellman Operator} in a value-iteration setting. The termination
gradient theorem can be interpreted as providing a gradient-based interrupting Bellman operator.

\section{Algorithms and Architecture}
\label{sect:arch}
\begin{figure}[ht!]
  \centering
  \includegraphics[width=0.75\columnwidth]{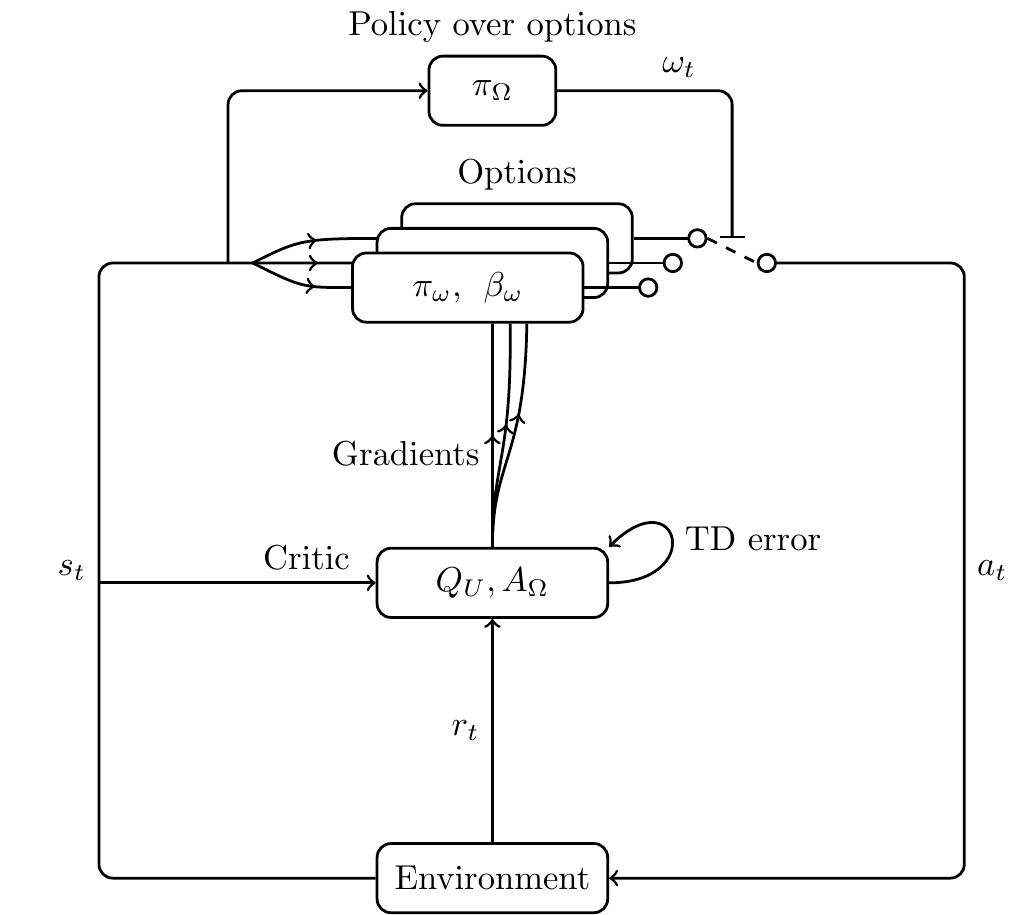}
  \caption{Diagram of the option-critic architecture.
  The option execution model is depicted by a \textit{switch} $\bot$ over the \textit{contacts} $\multimap$.
  A new option is selected according to $\pi_\Omega$ only when the current option terminates.}
  \label{fig:option-critic}
\end{figure}
Based on theorems \ref{thm:intrapg} and \ref{thm:term}, we can now design a
stochastic gradient descent algorithm for learning options.
Using a two-timescale framework~\cite{Konda1999},
we propose to learn the values at a \textit{fast} timescale while updating the
intra-option policies and termination functions at a \textit{slower} rate.

We refer to the resulting system as an \textit{option-critic architecture},
in reference to the actor-critic architectures \cite{Sutton1984}.
The intra-option policies, termination functions and
policy over options belong to the \textit{actor} part of the system while
the \textit{critic} consists of $Q_U$ and $A_{\Omega}$.
The option-critic architecture does not prescribe how to obtain $\pi_\Omega$ since
a variety of existing approaches would apply: using policy gradient
methods at the SMDP level, with a planner over the options models,
or using temporal difference updates. If $\pi_\Omega$ is the greedy
policy over options, it follows from \eqref{eq:qswa} that the corresponding
one-step off-policy update target $g_t^{(1)}$ is:
\begin{align*}
&g_t^{(1)} = r_{t+1} +  \\
  &\gamma \Big((1 - \beta_{\option_t, \termparams}(s_{t+1}))\sum_{a} \
  \pi_{\option_t, \theta}\left(a \given s_{t+1} \right)Q_U(s_{t+1}, \option_t, a)\notag\\
  & +  \beta_{\option_t, \termparams}(s_{t+1}) \max_{\option} \sum_{a} \pi_{\option, \theta}\left(a \given s_{t+1} \right)Q_U(s_{t+1}, \option, a) \Big)\enspace ,
\end{align*}
which is also the update target of the \textit{intra-option Q-learning} algorithm
of \cite{Sutton1999opt}.
A prototypical implementation of option-critic which uses intra-option Q-learning
is shown in Algorithm \ref{alg:oc-intraq}. The tabular setting is assumed only
for clarity of presentation.
We write $\alpha, \alpha_{\theta}$ and
$\alpha_{\vartheta}$ for the learning rates of the
critic, intra-option policies and termination functions respectively.

\begin{algorithm}[h]
\DontPrintSemicolon
$s \leftarrow s_0$ \;
Choose $\option$ according to an $\epsilon\text{-soft}$ policy over options $\pi_\Omega(s)$\;

\Repeat{$s'$ is terminal} {
Choose $a$ according to $\pi_{\option, \theta}\left( a \given s \right)$\;
Take action $a$ in $s$, observe $s'$, $r$\;
\;
\textbf{1. Options evaluation:}\;
$\delta \leftarrow r - Q_U(s, \option, a)$\;
\If{$s'$ is non-terminal}
{$\delta \leftarrow \delta + \gamma (1 - \beta_{\option, \termparams}(s'))Q_\options(s', \option) +
\gamma \beta_{\option, \termparams}(s') \max\limits_{\bar \option} Q_\options(s', \bar \option)$\;}
$Q_U(s, \option, a) \leftarrow Q_U(s, \option, a) + \alpha \delta$\;
\;
\textbf{2. Options improvement:}\;
$\theta \leftarrow \theta + \alpha_{\theta} \deriv[\log \pi_{\omega, \theta}\left( a \given s\right)]{\theta} Q_U(s, \option, a)$\;
$\vartheta \leftarrow \vartheta - \alpha_{\vartheta} \deriv[\beta_{\option, \vartheta}(s')]{\vartheta} \left( Q_\options(s', \option) - V_\options(s') \right)\;$\;
\;
\lIf{$\beta_{\omega, \vartheta}$ terminates in $s'$}{\;
choose new $\omega$ according to $\epsilon\text{-soft}(\pi_\Omega(s'))$ }
$s \leftarrow s'$\;
}

\caption{Option-critic with tabular intra-option Q-learning}
\label{alg:oc-intraq}
\end{algorithm}

Learning  $Q_U$ in addition to $Q_\Omega$ is computationally wasteful both
in terms of the number of parameters and samples. A practical solution is to
only learn $Q_\Omega$ and  derive an estimate of $Q_U$ from it. Because $Q_U$ is
an expectation over next states, $Q_U(s, \option, a) = \expectation_{s' \sim \prob}\left[ r(s, a) + \gamma U(\omega, s')\given s, \option, a\right]$,
it follows that $g_t^{(1)}$ is an appropriate estimator. We chose
this approach for our experiment with deep neural networks in the Arcade Learning Environment.

 \section{Experiments}\label{sect:experiments}

We first consider a navigation task in the four-rooms domain~\cite{Sutton1999opt}.
Our goal is to  evaluate the ability of a set of options learned fully autonomously
 to recover from a sudden change in the environment. \cite{Sutton1999opt} presented a similar
experiment for a set of pre-specified options; the options in our results
have not been specified a priori.

Initially the goal is located in the east doorway and the initial state
is drawn uniformly from all the other cells. After 1000 episodes,
the goal   moves to a random location in the lower right room. Primitive movements can fail with
probability $1/3$, in which case the agent transitions randomly to one of the
 empty adjacent cells. The discount factor was  $0.99$, and the reward was
$+1$  at the goal  and $0$ otherwise.
We chose to parametrize the intra-option policies with Boltzmann distributions
and the terminations with sigmoid functions. The policy
over options was learned using intra-option Q-learning.
We also implemented
primitive actor-critic  (denoted \textsc{AC-PG}) using a Boltzmann policy.
We also compared option-critic to a
primitive SARSA agent using Boltzmann exploration
and no eligibility traces.
For all Boltzmann policies, we set the temperature parameter to $0.001$.
All the weights were initialized to zero.

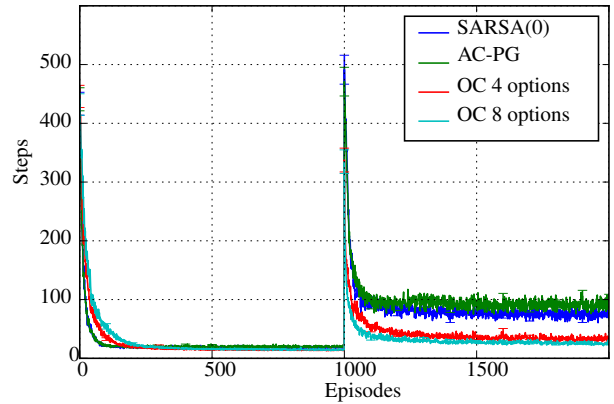
\begin{figure}[ht!]
  \centering
  \begin{subfigure}[b]{0.95\columnwidth}
  \def\svgwidth{\textwidth}
  \begingroup  \makeatletter  \providecommand\color[2][]{    \errmessage{(Inkscape) Color is used for the text in Inkscape, but the package 'color.sty' is not loaded}    \renewcommand\color[2][]{}  }  \providecommand\transparent[1]{    \errmessage{(Inkscape) Transparency is used (non-zero) for the text in Inkscape, but the package 'transparent.sty' is not loaded}    \renewcommand\transparent[1]{}  }  \providecommand\rotatebox[2]{#2}  \ifx\svgwidth\undefined    \setlength{\unitlength}{380.41391602bp}    \ifx\svgscale\undefined      \relax    \else      \setlength{\unitlength}{\unitlength * \real{\svgscale}}    \fi  \else    \setlength{\unitlength}{\svgwidth}  \fi  \global\let\svgwidth\undefined  \global\let\svgscale\undefined  \makeatother  {\fontsize{8pt}{1em} \selectfont  \begin{picture}(1,0.67470902)    \put(0,0){\includegraphics[width=\unitlength]{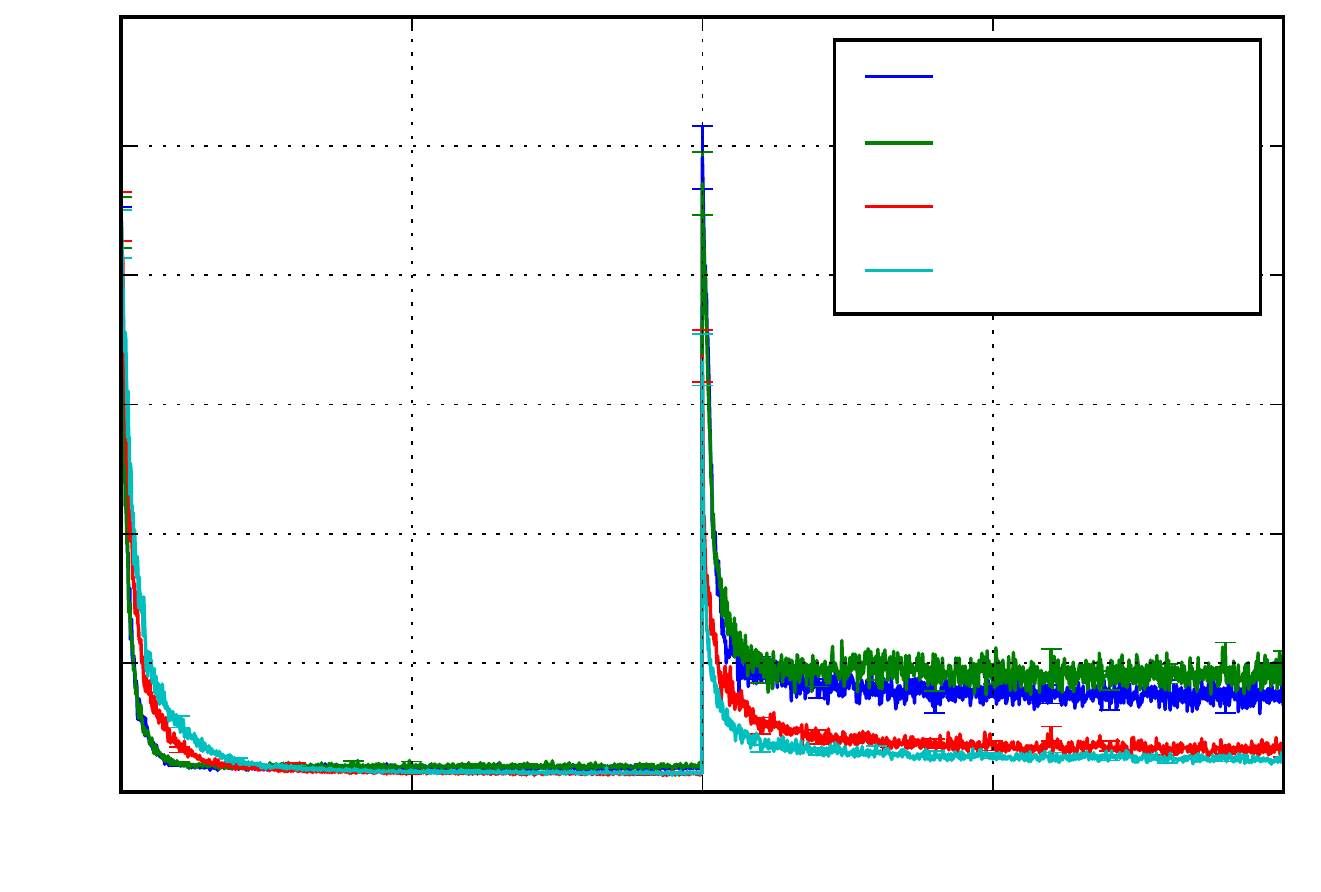}}    \put(0.025,0.1674864){\color[rgb]{0,0,0}\makebox(0,0)[lb]{\smash{100}}}    \put(0.025,0.26527462){\color[rgb]{0,0,0}\makebox(0,0)[lb]{\smash{200}}}    \put(0.025,0.36306284){\color[rgb]{0,0,0}\makebox(0,0)[lb]{\smash{300}}}    \put(0.025,0.46075855){\color[rgb]{0,0,0}\makebox(0,0)[lb]{\smash{400}}}    \put(0.025,0.5586393){\color[rgb]{0,0,0}\makebox(0,0)[lb]{\smash{500}}}    \put(0.07404504,0.07002677){\color[rgb]{0,0,0}\makebox(0,0)[lb]{\smash{0}}}    \put(0.08782945,0.04559328){\color[rgb]{0,0,0}\makebox(0,0)[lb]{\smash{0}}}    \put(0.29901993,0.04559328){\color[rgb]{0,0,0}\makebox(0,0)[lb]{\smash{500}}}    \put(0.51450943,0.04559328){\color[rgb]{0,0,0}\makebox(0,0)[lb]{\smash{1000}}}    \put(0.73453292,0.04559328){\color[rgb]{0,0,0}\makebox(0,0)[lb]{\smash{1500}}}    \put(0.49952363,0.01){\color[rgb]{0,0,0}\makebox(0,0)[lb]{\smash{Episodes}}}    \put(0.00459235,0.36418161){\rotatebox{90}{\smash{Steps}}}    \put(0.71926987,0.61467134){\color[rgb]{0,0,0}\makebox(0,0)[lb]{\smash{SARSA(0)}}}    \put(0.71967363,0.57035782){\color[rgb]{0,0,0}\makebox(0,0)[lb]{\smash{AC-PG}}}    \put(0.71926987,0.51889699){\color[rgb]{0,0,0}\makebox(0,0)[lb]{\smash{OC 4 options}}}    \put(0.71926987,0.47029507){\color[rgb]{0,0,0}\makebox(0,0)[lb]{\smash{OC 8 options}}}  \end{picture}}\endgroup   \end{subfigure}
   \caption{
   After a 1000 episodes, the goal location in the four-rooms domain
   is moved randomly. Option-critic (``OC'') recovers faster than the
   primitive actor-critic (``AC-PG'') and SARSA(0). Each line is averaged
   over 350 runs.
   }
   \label{fig:fourroomslr}
 \end{figure}

As can be seen in Figure \ref{fig:fourroomslr}, when the goal suddenly changes,
the option-critic agent recovers faster.
Furthermore, the initial set of options is learned \textit{from scratch} at a rate
comparable to  primitive methods. Despite the simplicity of the domain, we are not
aware of other methods which could have solved this task without incurring a cost
much larger than when using primitive actions  alone \cite{McGovern2001,Simsek08}.
\begin{figure}[ht!]
  \centering
  \includegraphics[width=0.5\columnwidth]{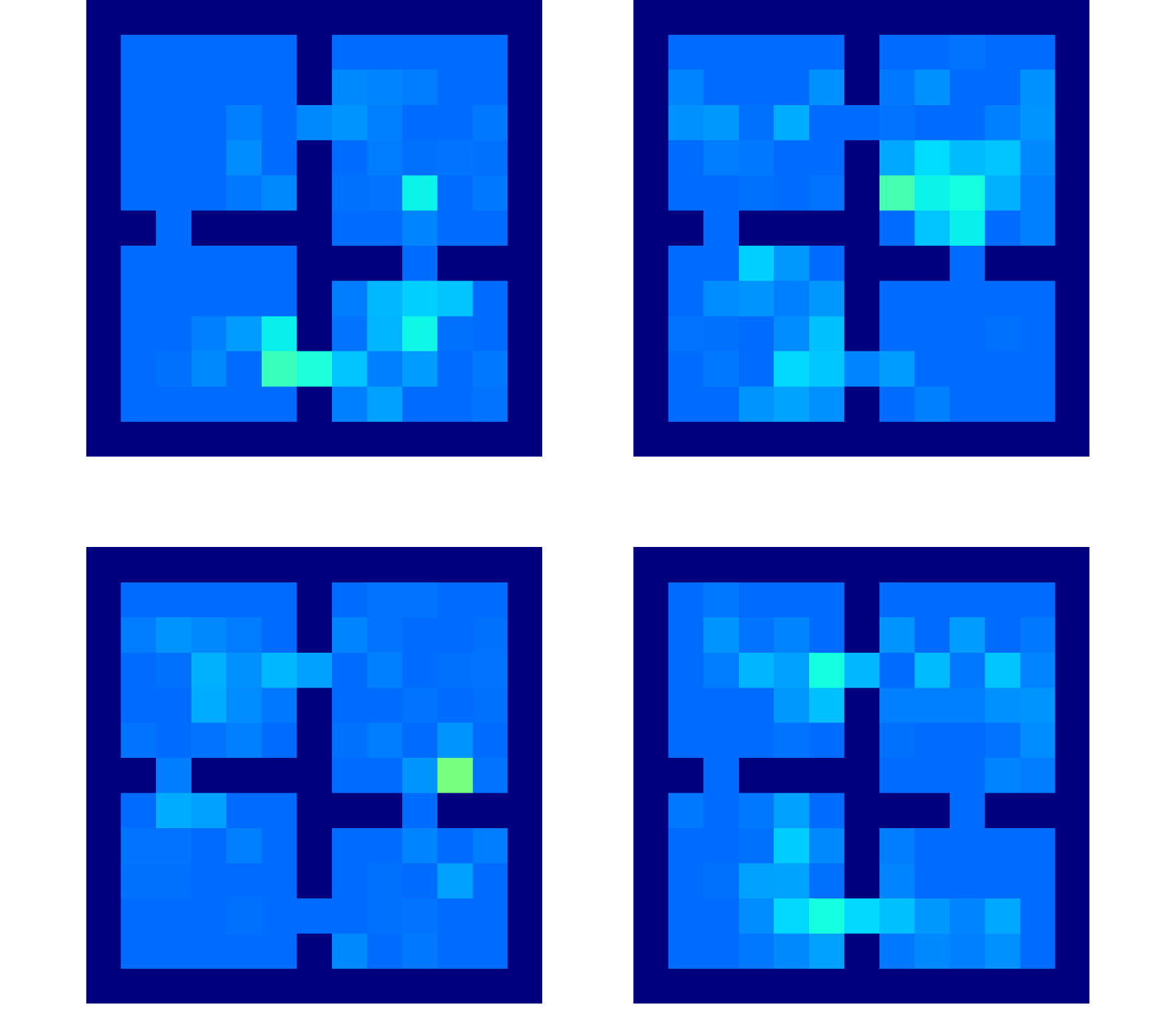}
  \caption{Termination probabilities for the option-critic agent learning with 4 options.
  The darkest color represents the
   \textit{walls} in the environment while lighter colors encode higher termination
  probabilities.}
  \label{fig:term}
\end{figure}

In the two temporally extended settings, with 4 options and 8 options,
termination events are more likely to occur near the doorways (Figure \ref{fig:term}), agreeing with the
intuition that they would be good subgoals.
As opposed to \cite{Sutton1999opt}, we did not encode this knowledge ourselves
but simply let the agents find options that would maximize the expected
discounted return.

\subsection{Pinball Domain}
\begin{figure}[ht!]
\centering
\includegraphics[width=0.4\columnwidth]{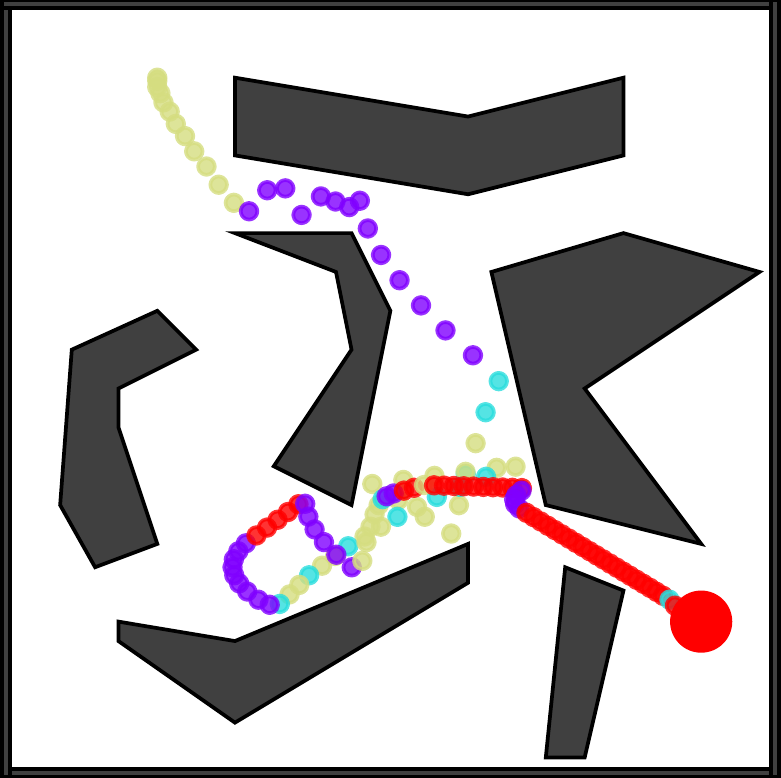}
\caption{Pinball: Sample trajectory of the
  solution found after 250 episodes of training using 4 options
  All options (color-coded) are used by the policy over options in successful trajectories. The initial state
  is in the top left corner and the goal is in the bottom right one (red circle). }
   \label{fig:pinballspecialization}
\end{figure}

In the Pinball domain \cite{Konidaris2009}, a ball must be guided through
a maze of arbitrarily shaped polygons to a designated target location. The
state space is continuous over the position and velocity of the
ball in the $x$-$y$ plane. At every step, the agent
must choose among five discrete primitive actions: move the ball faster or slower, in
the vertical or horizontal direction, or take the \textit{null} action.
Collisions with obstacles are elastic and can be used to the advantage of the agent.
In this domain, a drag coefficient of $0.995$ effectively stops ball movements after a finite number of steps when the null
action is chosen repeatedly. Each thrust action incurs a penalty of $-5$ while
taking no action costs $-1$. The episode terminates with $+10000$ reward when
the agent reaches the target. We interrupted any episode taking more than
$10000$ steps and set the discount factor to $0.99$.

We used intra-option Q-learning in the critic with linear function approximation
over Fourier bases~\cite{Konidaris2011} of order 3. We experimented with
2, 3 or 4 options. We used Boltzmann policies for the intra-option policies and linear-sigmoid
 functions for the termination functions.
 The learning rates were set to $0.01$
 for the critic and $0.001$ for both the intra and termination gradients. We used
 an epsilon-greedy policy over options with $\epsilon = 0.01$.

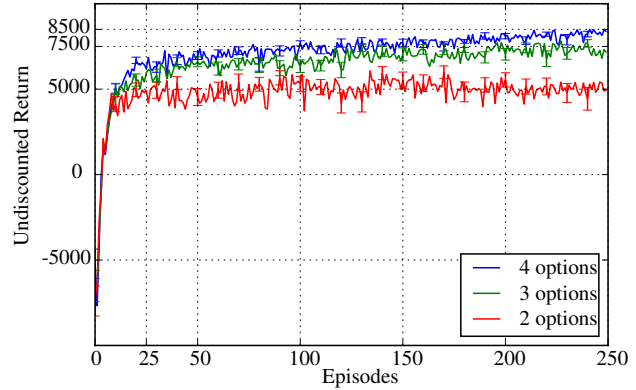
\begin{figure}[ht!]
\centering
\def\svgwidth{0.95\columnwidth}
\begingroup%
  \makeatletter%
  \providecommand\color[2][]{%
    \errmessage{(Inkscape) Color is used for the text in Inkscape, but the package 'color.sty' is not loaded}%
    \renewcommand\color[2][]{}%
  }%
  \providecommand\transparent[1]{%
    \errmessage{(Inkscape) Transparency is used (non-zero) for the text in Inkscape, but the package 'transparent.sty' is not loaded}%
    \renewcommand\transparent[1]{}%
  }%
  \providecommand\rotatebox[2]{#2}%
  \ifx\svgwidth\undefined%
    \setlength{\unitlength}{505.88999023bp}%
    \ifx\svgscale\undefined%
      \relax%
    \else%
      \setlength{\unitlength}{\unitlength * \real{\svgscale}}%
    \fi%
  \else%
    \setlength{\unitlength}{\svgwidth}%
  \fi%
  \global\let\svgwidth\undefined%
  \global\let\svgscale\undefined%
  \makeatother%
  {\fontsize{8pt}{1em} \selectfont%
  \begin{picture}(1,0.64725035)%
    \put(0,0){\includegraphics[width=\unitlength]{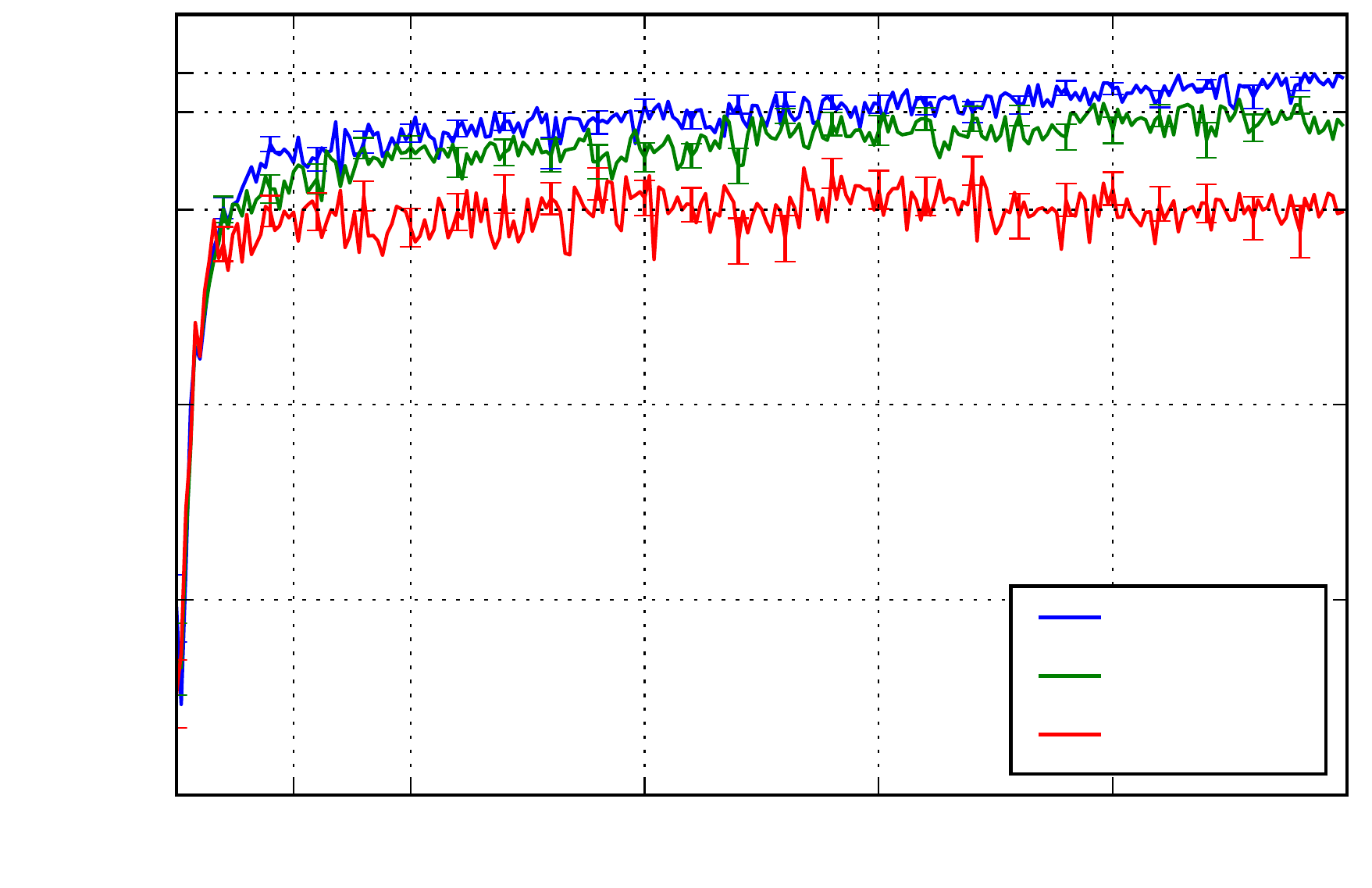}}%
    \put(0.12237095,0.035){\makebox(0,0)[lb]{\smash{0}}}%
    \put(0.20134882,0.035){\makebox(0,0)[lb]{\smash{25}}}%
    \put(0.28667426,0.035){\makebox(0,0)[lb]{\smash{50}}}%
    \put(0.45097756,0.035){\makebox(0,0)[lb]{\smash{100}}}%
    \put(0.62162844,0.035){\makebox(0,0)[lb]{\smash{150}}}%
    \put(0.79227932,0.035){\makebox(0,0)[lb]{\smash{200}}}%
    \put(0.96293019,0.035){\makebox(0,0)[lb]{\smash{250}}}%
    \put(0.50671633,0.00492569){\makebox(0,0)[lb]{\smash{Episodes}}}%
    \put(0.03530056,0.06150416){\makebox(0,0)[lb]{\smash{−}}}%
    \put(0.0479957,0.20371322){\makebox(0,0)[lb]{\smash{−}}}%
    \put(0.04,0.20371322){\makebox(0,0)[lb]{\smash{-5000}}}%
    \put(0.09,0.3459223){\makebox(0,0)[lb]{\smash{0}}}%
    \put(0.05,0.48813136){\makebox(0,0)[lb]{\smash{5000}}}%
    \put(0.05,0.5592359){\makebox(0,0)[lb]{\smash{7500}}}%
    \put(0.05,0.58767771){\makebox(0,0)[lb]{\smash{8500}}}%
    \put(0.01762086,0.23342921){\rotatebox{90}{\makebox(0,0)[lb]{\smash{Undiscounted Return}}}}%
    \put(0.83494785,0.18647665){\makebox(0,0)[lb]{\smash{4 options}}}%
    \put(0.83494785,0.14385319){\makebox(0,0)[lb]{\smash{3 options}}}%
    \put(0.83494785,0.10122973){\makebox(0,0)[lb]{\smash{2 options}}}%
  \end{picture}}%
\endgroup\caption{Learning curves in the Pinball domain.}
\end{figure}

In~\cite{Konidaris2009}, an option can only be used and updated
after a \textit{gestation} period of 10 episodes.
As learning is fully integrated in option-critic, by 40 episodes a near optimal
set of options had already been learned in all settings.
From a qualitative point
of view, the options exhibit temporal extension and specialization
(fig. \ref{fig:pinballspecialization}). We  also observed that
across many successful trajectories the \textit{red} option would consistently
be used in the vicinity of the goal.

\subsection{Arcade Learning Environment}
\label{sect:deep}

\begin{figure}[ht!]
  \centering{
  \def\svgwidth{0.9\columnwidth}
  \begingroup  \makeatletter  \providecommand\color[2][]{    \errmessage{(Inkscape) Color is used for the text in Inkscape, but the package 'color.sty' is not loaded}    \renewcommand\color[2][]{}  }  \providecommand\transparent[1]{    \errmessage{(Inkscape) Transparency is used (non-zero) for the text in Inkscape, but the package 'transparent.sty' is not loaded}    \renewcommand\transparent[1]{}  }  \providecommand\rotatebox[2]{#2}  \ifx\svgwidth\undefined    \setlength{\unitlength}{505.85297852bp}    \ifx\svgscale\undefined      \relax    \else      \setlength{\unitlength}{\unitlength * \real{\svgscale}}    \fi  \else    \setlength{\unitlength}{\svgwidth}  \fi  \global\let\svgwidth\undefined  \global\let\svgscale\undefined  \makeatother  {\fontsize{8pt}{1em} \selectfont  \begin{picture}(1,0.25428654)    \put(0,0){\includegraphics[width=\unitlength]{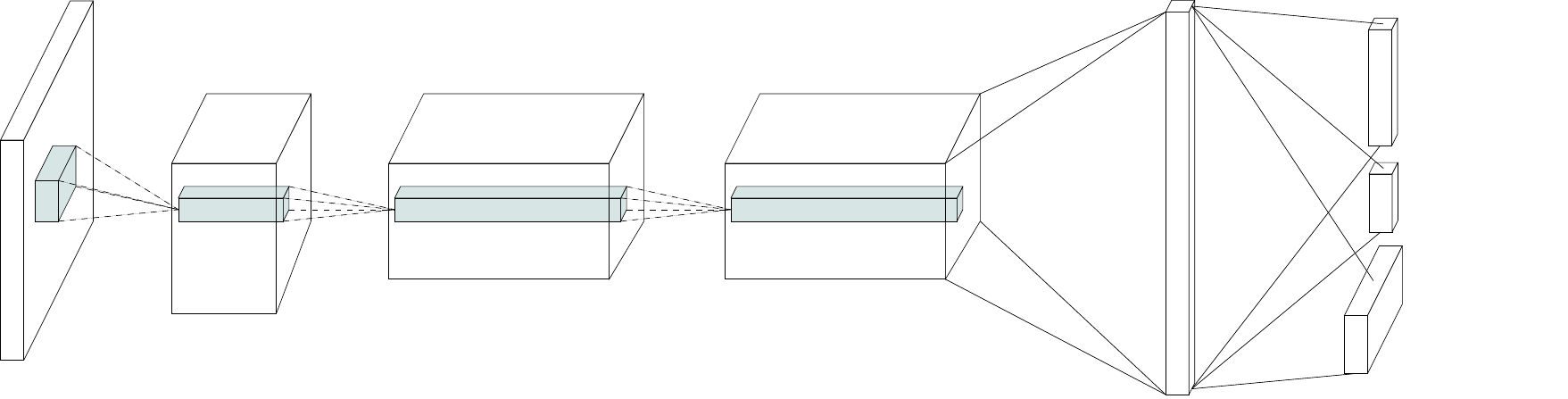}}    \put(0.90052662,0.19755272){\color[rgb]{0,0,0}\makebox(0,0)[lb]{\smash{$\pi_\Omega(\cdot | s)$}}}    \put(0.90052662,0.12153839){\color[rgb]{0,0,0}\makebox(0,0)[lb]{\smash{$\{\beta_\omega(s)\}$}}}    \put(0.90052662,0.05976149){\color[rgb]{0,0,0}\makebox(0,0)[lb]{\smash{$\{\pi_\omega(\cdot | s)\}$}}}  \end{picture}}\endgroup   \caption{Deep neural network architecture. A concatenation of the last
  4 images is fed through the convolutional layers, producing a dense
  representation shared across intra-option policies, termination functions
  and policy over options.}
  \label{fig:dnn-arch}}
\end{figure}
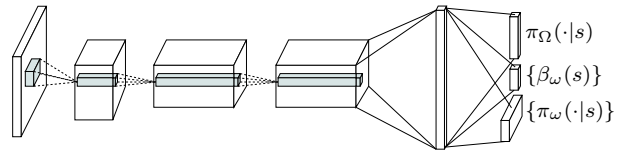

We applied the option-critic architecture in the Arcade Learning Environment
(ALE) \cite{bellemare13arcade} using a deep neural network
to approximate the critic and represent the intra-option policies and termination functions.
We used the same configuration as ~\cite{Mnih2013} for the first 3 convolutional
layers of the network.
We used $32$ convolutional filters of size $8 \times 8$
and stride of $4$ in the first layer, $64$ filters of size $4 \times 4$
with a stride of $2$ in the second and $64$ $3 \times 3$ filters with a
stride of $1$ in the third layer.
We then fed the output of the third layer
into a dense shared layer of $512$ neurons, as depicted in Figure~\ref{fig:dnn-arch}.
We fixed the learning rate for the
intra-option policies and termination gradient to $0.00025$ and used RMSProp for the critic.

We represented the intra-option policies as linear-softmax of the fourth
(dense) layer, so as to output a probability distribution over
actions conditioned on the current observation. The termination functions were similarly
defined using sigmoid functions, with one output neuron per termination.

The critic network was trained using intra-option Q-learning
with experience replay. Option policies and terminations were updated on-line. We used an
$\epsilon$-greedy policy over options with $\epsilon=0.05$ during the
\textit{test} phase~\cite{Mnih2013}.

As a consequence of optimizing for the return, the termination gradient
tends to \textit{shrink} options over time. This is expected since in theory
primitive actions are sufficient for solving any  MDP. We tackled this
issue by adding a small $\xi=0.01$ term to the advantage function,
used by the termination gradient: $A_\Omega(s, \omega) + \xi = Q_\Omega(s, \omega) - V_\Omega(s) + \xi$.
This term has a regularization effect, by imposing
an $\xi$-margin between the value estimate of an option and that of
the ``optimal'' one reflected in $V_\Omega$. This makes the advantage function
positive if the value of an option is near the optimal one, thereby
\textit{stretching} it. A similar regularizer was proposed in \cite{Mann2014}.

As in \cite{Mnih2016A3C}, we observed that the
intra-option policies would quickly become deterministic. This problem seems
to pertain to the use of policy gradient methods with deep neural networks in
general, and not from option-critic itself.
We applied the regularizer prescribed by \cite{Mnih2016A3C}, by penalizing for low-entropy
intra-option policies.

\begin{figure}[ht!]
  \centering
  \def\svgwidth{0.9\columnwidth}
  \begingroup%
    \makeatletter%
    \providecommand\color[2][]{%
      \errmessage{(Inkscape) Color is used for the text in Inkscape, but the package 'color.sty' is not loaded}%
      \renewcommand\color[2][]{}%
    }%
    \providecommand\transparent[1]{%
      \errmessage{(Inkscape) Transparency is used (non-zero) for the text in Inkscape, but the package 'transparent.sty' is not loaded}%
      \renewcommand\transparent[1]{}%
    }%
    \providecommand\rotatebox[2]{#2}%
    \ifx\svgwidth\undefined%
      \setlength{\unitlength}{444.24038086bp}%
      \ifx\svgscale\undefined%
        \relax%
      \else%
        \setlength{\unitlength}{\unitlength * \real{\svgscale}}%
      \fi%
    \else%
      \setlength{\unitlength}{\svgwidth}%
    \fi%
    \global\let\svgwidth\undefined%
    \global\let\svgscale\undefined%
    \makeatother%
    \begin{picture}(1,0.86087788)%
    {\fontsize{4pt}{0em} \selectfont%
      \put(0,0){\includegraphics[width=\unitlength]{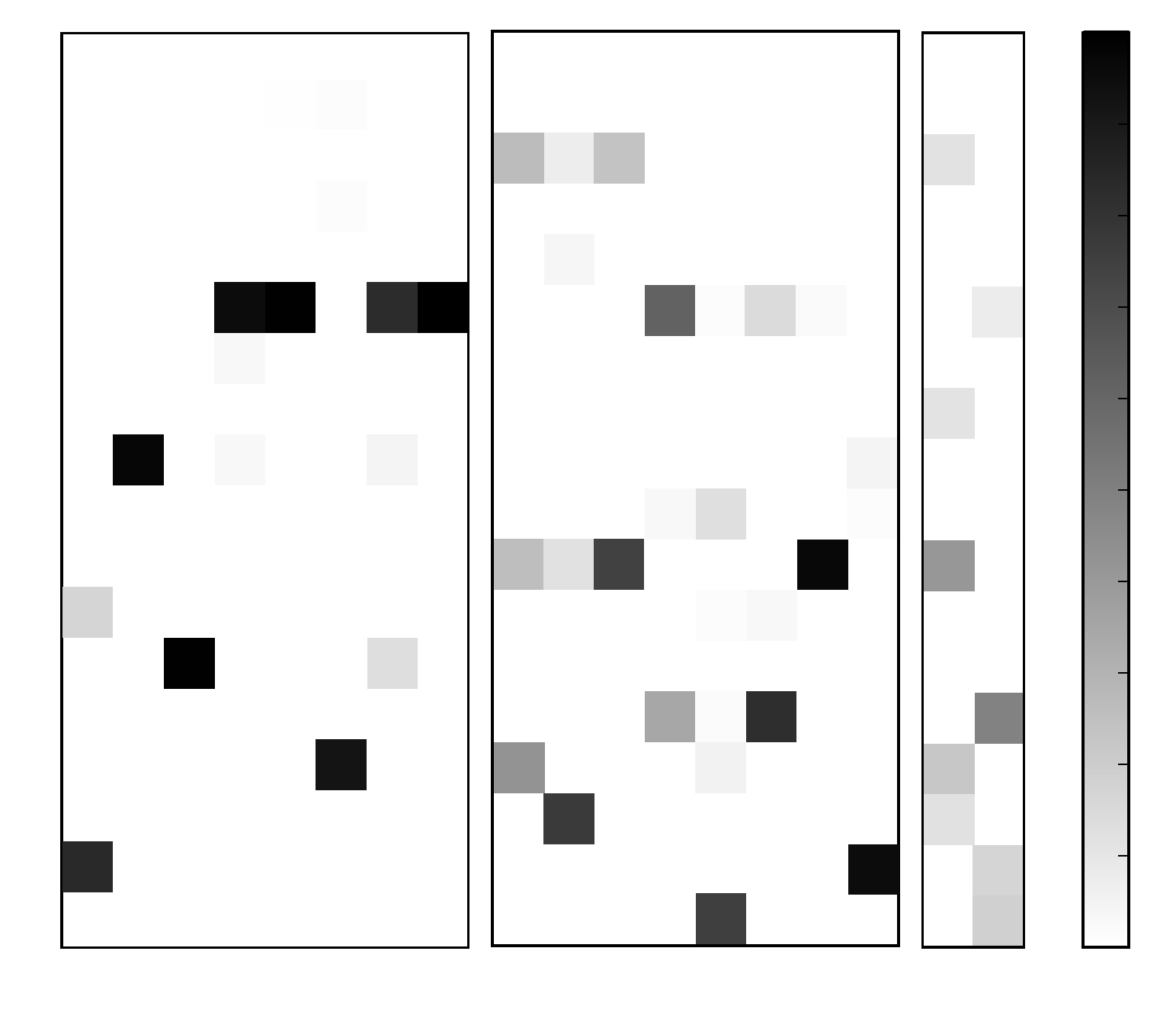}}%
      \put(0.06819875,0.02848831){\makebox(0,0)[lb]{\smash{1}}}%
      \put(0.1114186,0.02848831){\makebox(0,0)[lb]{\smash{2}}}%
      \put(0.15463845,0.02848831){\makebox(0,0)[lb]{\smash{3}}}%
      \put(0.19785828,0.02848831){\makebox(0,0)[lb]{\smash{4}}}%
      \put(0.24107812,0.02848831){\makebox(0,0)[lb]{\smash{5}}}%
      \put(0.28429797,0.02848831){\makebox(0,0)[lb]{\smash{6}}}%
      \put(0.32751781,0.02848831){\makebox(0,0)[lb]{\smash{7}}}%
      \put(0.37073765,0.02848831){\makebox(0,0)[lb]{\smash{8}}}%
      \put(0.03154598,0.80485568){\makebox(0,0)[lb]{\smash{0}}}%
      \put(0.03154598,0.76163588){\makebox(0,0)[lb]{\smash{1}}}%
      \put(0.03154598,0.718416){\makebox(0,0)[lb]{\smash{2}}}%
      \put(0.03154598,0.67519613){\makebox(0,0)[lb]{\smash{3}}}%
      \put(0.03154598,0.63197633){\makebox(0,0)[lb]{\smash{4}}}%
      \put(0.03154598,0.58875646){\makebox(0,0)[lb]{\smash{5}}}%
      \put(0.03154598,0.54553662){\makebox(0,0)[lb]{\smash{6}}}%
      \put(0.03154598,0.50231678){\makebox(0,0)[lb]{\smash{7}}}%
      \put(0.03154598,0.45909695){\makebox(0,0)[lb]{\smash{8}}}%
      \put(0.03154598,0.41587708){\makebox(0,0)[lb]{\smash{9}}}%
      \put(0.01946853,0.37265724){\makebox(0,0)[lb]{\smash{10}}}%
      \put(0.01946853,0.3294374){\makebox(0,0)[lb]{\smash{11}}}%
      \put(0.01946853,0.28621756){\makebox(0,0)[lb]{\smash{12}}}%
      \put(0.01946853,0.24299771){\makebox(0,0)[lb]{\smash{13}}}%
      \put(0.01946853,0.19977789){\makebox(0,0)[lb]{\smash{14}}}%
      \put(0.01946853,0.15655803){\makebox(0,0)[lb]{\smash{15}}}%
      \put(0.01946853,0.11333819){\makebox(0,0)[lb]{\smash{16}}}%
      \put(0.01946853,0.07011835){\makebox(0,0)[lb]{\smash{17}}}%
      \put(0.06905141,0.80608513){\makebox(0,0)[lb]{\smash{0}}}%
      \put(0.06905141,0.76286533){\makebox(0,0)[lb]{\smash{0}}}%
      \put(0.06905141,0.71964546){\makebox(0,0)[lb]{\smash{0}}}%
      \put(0.06905141,0.67642566){\makebox(0,0)[lb]{\smash{0}}}%
      \put(0.06905141,0.63320578){\makebox(0,0)[lb]{\smash{0}}}%
      \put(0.06905141,0.58998591){\makebox(0,0)[lb]{\smash{0}}}%
      \put(0.06905141,0.54676611){\makebox(0,0)[lb]{\smash{0}}}%
      \put(0.06905141,0.50354624){\makebox(0,0)[lb]{\smash{0}}}%
      \put(0.06905141,0.4603264){\makebox(0,0)[lb]{\smash{0}}}%
      \put(0.06905141,0.41710656){\makebox(0,0)[lb]{\smash{0}}}%
      \put(0.06905141,0.37388673){\makebox(0,0)[lb]{\smash{0}}}%
      \put(0.05579806,0.33066689){\makebox(0,0)[lb]{\smash{16.6}}}%
      \put(0.06905141,0.28744703){\makebox(0,0)[lb]{\smash{0}}}%
      \put(0.06905141,0.24422718){\makebox(0,0)[lb]{\smash{0}}}%
      \put(0.06905141,0.20100734){\makebox(0,0)[lb]{\smash{0}}}%
      \put(0.06905141,0.15778751){\makebox(0,0)[lb]{\smash{0}}}%
      \put(0.05579806,0.11456765){\color{white}\makebox(0,0)[lb]{\smash{83.4}}}%
      \put(0.06905141,0.07134781){\makebox(0,0)[lb]{\smash{0}}}%
      \put(0.10420396,0.80608513){\makebox(0,0)[lb]{\smash{0.2}}}%
      \put(0.11227125,0.76286533){\makebox(0,0)[lb]{\smash{0}}}%
      \put(0.10420396,0.71964546){\makebox(0,0)[lb]{\smash{0.2}}}%
      \put(0.11227125,0.67642566){\makebox(0,0)[lb]{\smash{0}}}%
      \put(0.11227125,0.63320578){\makebox(0,0)[lb]{\smash{0}}}%
      \put(0.11227125,0.58998591){\makebox(0,0)[lb]{\smash{0}}}%
      \put(0.11227125,0.54676611){\makebox(0,0)[lb]{\smash{0}}}%
      \put(0.10420396,0.50354624){\makebox(0,0)[lb]{\smash{0.2}}}%
      \put(0.0990179,0.4603264){\color{white}\makebox(0,0)[lb]{\smash{97.8}}}%
      \put(0.11227125,0.41710656){\makebox(0,0)[lb]{\smash{0}}}%
      \put(0.11227125,0.37388673){\makebox(0,0)[lb]{\smash{0}}}%
      \put(0.11227125,0.33066689){\makebox(0,0)[lb]{\smash{0}}}%
      \put(0.11227125,0.28744703){\makebox(0,0)[lb]{\smash{0}}}%
      \put(0.11227125,0.24422718){\makebox(0,0)[lb]{\smash{0}}}%
      \put(0.11227125,0.20100734){\makebox(0,0)[lb]{\smash{0}}}%
      \put(0.11227125,0.15778751){\makebox(0,0)[lb]{\smash{0}}}%
      \put(0.10420396,0.11456765){\makebox(0,0)[lb]{\smash{1.5}}}%
      \put(0.11227125,0.07134781){\makebox(0,0)[lb]{\smash{0}}}%
      \put(0.1554911,0.80608513){\makebox(0,0)[lb]{\smash{0}}}%
      \put(0.1554911,0.76286533){\makebox(0,0)[lb]{\smash{0}}}%
      \put(0.1554911,0.71964546){\makebox(0,0)[lb]{\smash{0}}}%
      \put(0.1554911,0.67642566){\makebox(0,0)[lb]{\smash{0}}}%
      \put(0.1554911,0.63320578){\makebox(0,0)[lb]{\smash{0}}}%
      \put(0.1554911,0.58998591){\makebox(0,0)[lb]{\smash{0}}}%
      \put(0.1554911,0.54676611){\makebox(0,0)[lb]{\smash{0}}}%
      \put(0.1554911,0.50354624){\makebox(0,0)[lb]{\smash{0}}}%
      \put(0.1554911,0.4603264){\makebox(0,0)[lb]{\smash{0}}}%
      \put(0.1554911,0.41710656){\makebox(0,0)[lb]{\smash{0}}}%
      \put(0.1554911,0.37388673){\makebox(0,0)[lb]{\smash{0}}}%
      \put(0.1554911,0.33066689){\makebox(0,0)[lb]{\smash{0}}}%
      \put(0.14511896,0.28744703){\color{white}\makebox(0,0)[lb]{\smash{100}}}%
      \put(0.1554911,0.24422718){\makebox(0,0)[lb]{\smash{0}}}%
      \put(0.1554911,0.20100734){\makebox(0,0)[lb]{\smash{0}}}%
      \put(0.1554911,0.15778751){\makebox(0,0)[lb]{\smash{0}}}%
      \put(0.1554911,0.11456765){\makebox(0,0)[lb]{\smash{0}}}%
      \put(0.1554911,0.07134781){\makebox(0,0)[lb]{\smash{0}}}%
      \put(0.19871094,0.80608513){\makebox(0,0)[lb]{\smash{0}}}%
      \put(0.19871094,0.76286533){\makebox(0,0)[lb]{\smash{0}}}%
      \put(0.19871094,0.71964546){\makebox(0,0)[lb]{\smash{0}}}%
      \put(0.19871094,0.67642566){\makebox(0,0)[lb]{\smash{0}}}%
      \put(0.19871094,0.63320578){\makebox(0,0)[lb]{\smash{0}}}%
      \put(0.18545759,0.58998591){\color{white}\makebox(0,0)[lb]{\smash{95.2}}}%
      \put(0.19064365,0.54676611){\makebox(0,0)[lb]{\smash{2.4}}}%
      \put(0.19871094,0.50354624){\makebox(0,0)[lb]{\smash{0}}}%
      \put(0.19064365,0.4603264){\makebox(0,0)[lb]{\smash{2.4}}}%
      \put(0.19871094,0.41710656){\makebox(0,0)[lb]{\smash{0}}}%
      \put(0.19871094,0.37388673){\makebox(0,0)[lb]{\smash{0}}}%
      \put(0.19871094,0.33066689){\makebox(0,0)[lb]{\smash{0}}}%
      \put(0.19871094,0.28744703){\makebox(0,0)[lb]{\smash{0}}}%
      \put(0.19871094,0.24422718){\makebox(0,0)[lb]{\smash{0}}}%
      \put(0.19871094,0.20100734){\makebox(0,0)[lb]{\smash{0}}}%
      \put(0.19871094,0.15778751){\makebox(0,0)[lb]{\smash{0}}}%
      \put(0.19871094,0.11456765){\makebox(0,0)[lb]{\smash{0}}}%
      \put(0.19871094,0.07134781){\makebox(0,0)[lb]{\smash{0}}}%
      \put(0.24193078,0.80608513){\makebox(0,0)[lb]{\smash{0}}}%
      \put(0.24193078,0.76286533){\makebox(0,0)[lb]{\smash{0}}}%
      \put(0.24193078,0.71964546){\makebox(0,0)[lb]{\smash{0}}}%
      \put(0.24193078,0.67642566){\makebox(0,0)[lb]{\smash{0}}}%
      \put(0.24193078,0.63320578){\makebox(0,0)[lb]{\smash{0}}}%
      \put(0.23155864,0.58998591){\color{white}\makebox(0,0)[lb]{\smash{100}}}%
      \put(0.24193078,0.54676611){\makebox(0,0)[lb]{\smash{0}}}%
      \put(0.24193078,0.50354624){\makebox(0,0)[lb]{\smash{0}}}%
      \put(0.24193078,0.4603264){\makebox(0,0)[lb]{\smash{0}}}%
      \put(0.24193078,0.41710656){\makebox(0,0)[lb]{\smash{0}}}%
      \put(0.24193078,0.37388673){\makebox(0,0)[lb]{\smash{0}}}%
      \put(0.24193078,0.33066689){\makebox(0,0)[lb]{\smash{0}}}%
      \put(0.24193078,0.28744703){\makebox(0,0)[lb]{\smash{0}}}%
      \put(0.24193078,0.24422718){\makebox(0,0)[lb]{\smash{0}}}%
      \put(0.24193078,0.20100734){\makebox(0,0)[lb]{\smash{0}}}%
      \put(0.24193078,0.15778751){\makebox(0,0)[lb]{\smash{0}}}%
      \put(0.24193078,0.11456765){\makebox(0,0)[lb]{\smash{0}}}%
      \put(0.24193078,0.07134781){\makebox(0,0)[lb]{\smash{0}}}%
      \put(0.27708334,0.80608513){\makebox(0,0)[lb]{\smash{0.3}}}%
      \put(0.27708334,0.76286533){\makebox(0,0)[lb]{\smash{1.0}}}%
      \put(0.27708334,0.71964546){\makebox(0,0)[lb]{\smash{0.4}}}%
      \put(0.27708334,0.67642566){\makebox(0,0)[lb]{\smash{1.0}}}%
      \put(0.27708334,0.63320578){\makebox(0,0)[lb]{\smash{0.3}}}%
      \put(0.27708334,0.58998591){\makebox(0,0)[lb]{\smash{0.1}}}%
      \put(0.27708334,0.54676611){\makebox(0,0)[lb]{\smash{0.3}}}%
      \put(0.27708334,0.50354624){\makebox(0,0)[lb]{\smash{0.3}}}%
      \put(0.27708334,0.4603264){\makebox(0,0)[lb]{\smash{0.4}}}%
      \put(0.27708334,0.41710656){\makebox(0,0)[lb]{\smash{0.1}}}%
      \put(0.27708334,0.37388673){\makebox(0,0)[lb]{\smash{0.4}}}%
      \put(0.27708334,0.33066689){\makebox(0,0)[lb]{\smash{0.3}}}%
      \put(0.27708334,0.28744703){\makebox(0,0)[lb]{\smash{0.3}}}%
      \put(0.27708334,0.24422718){\makebox(0,0)[lb]{\smash{0.3}}}%
      \put(0.27189728,0.20100734){\color{white}\makebox(0,0)[lb]{\smash{91.9}}}%
      \put(0.28515061,0.15778751){\makebox(0,0)[lb]{\smash{0}}}%
      \put(0.27708334,0.11456765){\makebox(0,0)[lb]{\smash{1.5}}}%
      \put(0.27708334,0.07134781){\makebox(0,0)[lb]{\smash{1.2}}}%
      \put(0.32837047,0.80608513){\makebox(0,0)[lb]{\smash{0}}}%
      \put(0.32837047,0.76286533){\makebox(0,0)[lb]{\smash{0}}}%
      \put(0.32837047,0.71964546){\makebox(0,0)[lb]{\smash{0}}}%
      \put(0.32837047,0.67642566){\makebox(0,0)[lb]{\smash{0}}}%
      \put(0.32837047,0.63320578){\makebox(0,0)[lb]{\smash{0}}}%
      \put(0.31511714,0.58998591){\color{white}\makebox(0,0)[lb]{\smash{82.6}}}%
      \put(0.32837047,0.54676611){\makebox(0,0)[lb]{\smash{0}}}%
      \put(0.32837047,0.50354624){\makebox(0,0)[lb]{\smash{0}}}%
      \put(0.32030318,0.4603264){\makebox(0,0)[lb]{\smash{4.3}}}%
      \put(0.32837047,0.41710656){\makebox(0,0)[lb]{\smash{0}}}%
      \put(0.32837047,0.37388673){\makebox(0,0)[lb]{\smash{0}}}%
      \put(0.32837047,0.33066689){\makebox(0,0)[lb]{\smash{0}}}%
      \put(0.31511714,0.28744703){\makebox(0,0)[lb]{\smash{13.0}}}%
      \put(0.32837047,0.24422718){\makebox(0,0)[lb]{\smash{0}}}%
      \put(0.32837047,0.20100734){\makebox(0,0)[lb]{\smash{0}}}%
      \put(0.32837047,0.15778751){\makebox(0,0)[lb]{\smash{0}}}%
      \put(0.32837047,0.11456765){\makebox(0,0)[lb]{\smash{0}}}%
      \put(0.32837047,0.07134781){\makebox(0,0)[lb]{\smash{0}}}%
      \put(0.3715903,0.80608513){\makebox(0,0)[lb]{\smash{0}}}%
      \put(0.3715903,0.76286533){\makebox(0,0)[lb]{\smash{0}}}%
      \put(0.3715903,0.71964546){\makebox(0,0)[lb]{\smash{0}}}%
      \put(0.3715903,0.67642566){\makebox(0,0)[lb]{\smash{0}}}%
      \put(0.3715903,0.63320578){\makebox(0,0)[lb]{\smash{0}}}%
      \put(0.36121819,0.58998591){\color{white}\makebox(0,0)[lb]{\smash{100}}}%
      \put(0.3715903,0.54676611){\makebox(0,0)[lb]{\smash{0}}}%
      \put(0.3715903,0.50354624){\makebox(0,0)[lb]{\smash{0}}}%
      \put(0.3715903,0.4603264){\makebox(0,0)[lb]{\smash{0}}}%
      \put(0.3715903,0.41710656){\makebox(0,0)[lb]{\smash{0}}}%
      \put(0.3715903,0.37388673){\makebox(0,0)[lb]{\smash{0}}}%
      \put(0.3715903,0.33066689){\makebox(0,0)[lb]{\smash{0}}}%
      \put(0.3715903,0.28744703){\makebox(0,0)[lb]{\smash{0}}}%
      \put(0.3715903,0.24422718){\makebox(0,0)[lb]{\smash{0}}}%
      \put(0.3715903,0.20100734){\makebox(0,0)[lb]{\smash{0}}}%
      \put(0.3715903,0.15778751){\makebox(0,0)[lb]{\smash{0}}}%
      \put(0.3715903,0.11456765){\makebox(0,0)[lb]{\smash{0}}}%
      \put(0.3715903,0.07134781){\makebox(0,0)[lb]{\smash{0}}}%
      \put(0.43421679,0.03012841){\makebox(0,0)[lb]{\smash{1}}}%
      \put(0.47743666,0.03012841){\makebox(0,0)[lb]{\smash{2}}}%
      \put(0.52065646,0.03012841){\makebox(0,0)[lb]{\smash{3}}}%
      \put(0.56387633,0.03012841){\makebox(0,0)[lb]{\smash{4}}}%
      \put(0.60709621,0.03012841){\makebox(0,0)[lb]{\smash{5}}}%
      \put(0.65031601,0.03012841){\makebox(0,0)[lb]{\smash{6}}}%
      \put(0.69353588,0.03012841){\makebox(0,0)[lb]{\smash{7}}}%
      \put(0.73675568,0.03012841){\makebox(0,0)[lb]{\smash{8}}}%
      \put(0.42700219,0.80772522){\makebox(0,0)[lb]{\smash{1.1}}}%
      \put(0.42700219,0.76450542){\makebox(0,0)[lb]{\smash{1.1}}}%
      \put(0.42181611,0.72128554){\makebox(0,0)[lb]{\smash{26.1}}}%
      \put(0.43506944,0.67806574){\makebox(0,0)[lb]{\smash{0}}}%
      \put(0.42700219,0.63484587){\makebox(0,0)[lb]{\smash{1.1}}}%
      \put(0.43506944,0.59162603){\makebox(0,0)[lb]{\smash{0}}}%
      \put(0.42700219,0.5484062){\makebox(0,0)[lb]{\smash{1.1}}}%
      \put(0.43506944,0.50518636){\makebox(0,0)[lb]{\smash{0}}}%
      \put(0.43506944,0.46196649){\makebox(0,0)[lb]{\smash{0}}}%
      \put(0.43506944,0.41874665){\makebox(0,0)[lb]{\smash{0}}}%
      \put(0.42181611,0.37552681){\makebox(0,0)[lb]{\smash{25.0}}}%
      \put(0.43506944,0.33230697){\makebox(0,0)[lb]{\smash{0}}}%
      \put(0.43506944,0.28908712){\makebox(0,0)[lb]{\smash{0}}}%
      \put(0.43506944,0.24586728){\makebox(0,0)[lb]{\smash{0}}}%
      \put(0.42181611,0.20264745){\makebox(0,0)[lb]{\smash{42.0}}}%
      \put(0.43506944,0.15942759){\makebox(0,0)[lb]{\smash{0}}}%
      \put(0.42700219,0.11620775){\makebox(0,0)[lb]{\smash{2.3}}}%
      \put(0.43506944,0.07298791){\makebox(0,0)[lb]{\smash{0}}}%
      \put(0.47828932,0.80772522){\makebox(0,0)[lb]{\smash{0}}}%
      \put(0.47828932,0.76450542){\makebox(0,0)[lb]{\smash{0}}}%
      \put(0.47022199,0.72128554){\makebox(0,0)[lb]{\smash{7.0}}}%
      \put(0.47828932,0.67806574){\makebox(0,0)[lb]{\smash{0}}}%
      \put(0.47022199,0.63484587){\makebox(0,0)[lb]{\smash{3.8}}}%
      \put(0.47828932,0.59162603){\makebox(0,0)[lb]{\smash{0}}}%
      \put(0.47828932,0.5484062){\makebox(0,0)[lb]{\smash{0}}}%
      \put(0.47828932,0.50518636){\makebox(0,0)[lb]{\smash{0}}}%
      \put(0.47828932,0.46196649){\makebox(0,0)[lb]{\smash{0}}}%
      \put(0.47828932,0.41874665){\makebox(0,0)[lb]{\smash{0}}}%
      \put(0.46503598,0.37552681){\makebox(0,0)[lb]{\smash{11.4}}}%
      \put(0.47828932,0.33230697){\makebox(0,0)[lb]{\smash{0}}}%
      \put(0.47828932,0.28908712){\makebox(0,0)[lb]{\smash{0}}}%
      \put(0.47828932,0.24586728){\makebox(0,0)[lb]{\smash{0}}}%
      \put(0.47828932,0.20264745){\makebox(0,0)[lb]{\smash{0}}}%
      \put(0.46503598,0.15942759){\color{white}\makebox(0,0)[lb]{\smash{77.2}}}%
      \put(0.47828932,0.11620775){\makebox(0,0)[lb]{\smash{0}}}%
      \put(0.47022199,0.07298791){\makebox(0,0)[lb]{\smash{0.6}}}%
      \put(0.52150912,0.80772522){\makebox(0,0)[lb]{\smash{0}}}%
      \put(0.52150912,0.76450542){\makebox(0,0)[lb]{\smash{0}}}%
      \put(0.50825579,0.72128554){\makebox(0,0)[lb]{\smash{23.7}}}%
      \put(0.52150912,0.67806574){\makebox(0,0)[lb]{\smash{0}}}%
      \put(0.52150912,0.63484587){\makebox(0,0)[lb]{\smash{0}}}%
      \put(0.52150912,0.59162603){\makebox(0,0)[lb]{\smash{0}}}%
      \put(0.52150912,0.5484062){\makebox(0,0)[lb]{\smash{0}}}%
      \put(0.52150912,0.50518636){\makebox(0,0)[lb]{\smash{0}}}%
      \put(0.52150912,0.46196649){\makebox(0,0)[lb]{\smash{0}}}%
      \put(0.52150912,0.41874665){\makebox(0,0)[lb]{\smash{0}}}%
      \put(0.50825579,0.37552681){\color{white}\makebox(0,0)[lb]{\smash{74.8}}}%
      \put(0.52150912,0.33230697){\makebox(0,0)[lb]{\smash{0}}}%
      \put(0.52150912,0.28908712){\makebox(0,0)[lb]{\smash{0}}}%
      \put(0.52150912,0.24586728){\makebox(0,0)[lb]{\smash{0}}}%
      \put(0.52150912,0.20264745){\makebox(0,0)[lb]{\smash{0}}}%
      \put(0.52150912,0.15942759){\makebox(0,0)[lb]{\smash{0}}}%
      \put(0.51344186,0.11620775){\makebox(0,0)[lb]{\smash{1.5}}}%
      \put(0.52150912,0.07298791){\makebox(0,0)[lb]{\smash{0}}}%
      \put(0.56472899,0.80772522){\makebox(0,0)[lb]{\smash{0}}}%
      \put(0.56472899,0.76450542){\makebox(0,0)[lb]{\smash{0}}}%
      \put(0.55666173,0.72128554){\makebox(0,0)[lb]{\smash{0.3}}}%
      \put(0.56472899,0.67806574){\makebox(0,0)[lb]{\smash{0}}}%
      \put(0.56472899,0.63484587){\makebox(0,0)[lb]{\smash{0}}}%
      \put(0.55147566,0.59162603){\color{white}\makebox(0,0)[lb]{\smash{61.7}}}%
      \put(0.56472899,0.5484062){\makebox(0,0)[lb]{\smash{0}}}%
      \put(0.56472899,0.50518636){\makebox(0,0)[lb]{\smash{0}}}%
      \put(0.55666173,0.46196649){\makebox(0,0)[lb]{\smash{0.3}}}%
      \put(0.55666173,0.41874665){\makebox(0,0)[lb]{\smash{2.0}}}%
      \put(0.56472899,0.37552681){\makebox(0,0)[lb]{\smash{0}}}%
      \put(0.56472899,0.33230697){\makebox(0,0)[lb]{\smash{0}}}%
      \put(0.55666173,0.28908712){\makebox(0,0)[lb]{\smash{1.0}}}%
      \put(0.55147566,0.24586728){\makebox(0,0)[lb]{\smash{34.3}}}%
      \put(0.56472899,0.20264745){\makebox(0,0)[lb]{\smash{0}}}%
      \put(0.56472899,0.15942759){\makebox(0,0)[lb]{\smash{0}}}%
      \put(0.56472899,0.11620775){\makebox(0,0)[lb]{\smash{0}}}%
      \put(0.55666173,0.07298791){\makebox(0,0)[lb]{\smash{0.3}}}%
      \put(0.59988154,0.80772522){\makebox(0,0)[lb]{\smash{0.6}}}%
      \put(0.60794886,0.76450542){\makebox(0,0)[lb]{\smash{0}}}%
      \put(0.59988154,0.72128554){\makebox(0,0)[lb]{\smash{1.2}}}%
      \put(0.59988154,0.67806574){\makebox(0,0)[lb]{\smash{0.6}}}%
      \put(0.59988154,0.63484587){\makebox(0,0)[lb]{\smash{0.6}}}%
      \put(0.59988154,0.59162603){\makebox(0,0)[lb]{\smash{0.6}}}%
      \put(0.60794886,0.5484062){\makebox(0,0)[lb]{\smash{0}}}%
      \put(0.59988154,0.50518636){\makebox(0,0)[lb]{\smash{0.6}}}%
      \put(0.60794886,0.46196649){\makebox(0,0)[lb]{\smash{0}}}%
      \put(0.59469546,0.41874665){\makebox(0,0)[lb]{\smash{12.5}}}%
      \put(0.60794886,0.37552681){\makebox(0,0)[lb]{\smash{0}}}%
      \put(0.59988154,0.33230697){\makebox(0,0)[lb]{\smash{0.6}}}%
      \put(0.60794886,0.28908712){\makebox(0,0)[lb]{\smash{0}}}%
      \put(0.59988154,0.24586728){\makebox(0,0)[lb]{\smash{1.2}}}%
      \put(0.59988154,0.20264745){\makebox(0,0)[lb]{\smash{5.4}}}%
      \put(0.60794886,0.15942759){\makebox(0,0)[lb]{\smash{0}}}%
      \put(0.59988154,0.11620775){\makebox(0,0)[lb]{\smash{0.6}}}%
      \put(0.59469546,0.07298791){\color{white}\makebox(0,0)[lb]{\smash{75.6}}}%
      \put(0.65116866,0.80772522){\makebox(0,0)[lb]{\smash{0}}}%
      \put(0.65116866,0.76450542){\makebox(0,0)[lb]{\smash{0}}}%
      \put(0.65116866,0.72128554){\makebox(0,0)[lb]{\smash{0}}}%
      \put(0.65116866,0.67806574){\makebox(0,0)[lb]{\smash{0}}}%
      \put(0.65116866,0.63484587){\makebox(0,0)[lb]{\smash{0}}}%
      \put(0.63791533,0.59162603){\makebox(0,0)[lb]{\smash{14.1}}}%
      \put(0.65116866,0.5484062){\makebox(0,0)[lb]{\smash{0}}}%
      \put(0.65116866,0.50518636){\makebox(0,0)[lb]{\smash{0}}}%
      \put(0.65116866,0.46196649){\makebox(0,0)[lb]{\smash{0}}}%
      \put(0.65116866,0.41874665){\makebox(0,0)[lb]{\smash{0}}}%
      \put(0.65116866,0.37552681){\makebox(0,0)[lb]{\smash{0}}}%
      \put(0.64310141,0.33230697){\makebox(0,0)[lb]{\smash{2.4}}}%
      \put(0.65116866,0.28908712){\makebox(0,0)[lb]{\smash{0}}}%
      \put(0.63791533,0.24586728){\color{white}\makebox(0,0)[lb]{\smash{82.4}}}%
      \put(0.65116866,0.20264745){\makebox(0,0)[lb]{\smash{0}}}%
      \put(0.64310141,0.15942759){\makebox(0,0)[lb]{\smash{1.2}}}%
      \put(0.65116866,0.11620775){\makebox(0,0)[lb]{\smash{0}}}%
      \put(0.65116866,0.07298791){\makebox(0,0)[lb]{\smash{0}}}%
      \put(0.69438854,0.80772522){\makebox(0,0)[lb]{\smash{0}}}%
      \put(0.69438854,0.76450542){\makebox(0,0)[lb]{\smash{0}}}%
      \put(0.69438854,0.72128554){\makebox(0,0)[lb]{\smash{0}}}%
      \put(0.69438854,0.67806574){\makebox(0,0)[lb]{\smash{0}}}%
      \put(0.69438854,0.63484587){\makebox(0,0)[lb]{\smash{0}}}%
      \put(0.68632121,0.59162603){\makebox(0,0)[lb]{\smash{2.1}}}%
      \put(0.68632121,0.5484062){\makebox(0,0)[lb]{\smash{1.0}}}%
      \put(0.69438854,0.50518636){\makebox(0,0)[lb]{\smash{0}}}%
      \put(0.69438854,0.46196649){\makebox(0,0)[lb]{\smash{0}}}%
      \put(0.69438854,0.41874665){\makebox(0,0)[lb]{\smash{0}}}%
      \put(0.6811352,0.37552681){\color{white}\makebox(0,0)[lb]{\smash{96.9}}}%
      \put(0.69438854,0.33230697){\makebox(0,0)[lb]{\smash{0}}}%
      \put(0.69438854,0.28908712){\makebox(0,0)[lb]{\smash{0}}}%
      \put(0.69438854,0.24586728){\makebox(0,0)[lb]{\smash{0}}}%
      \put(0.69438854,0.20264745){\makebox(0,0)[lb]{\smash{0}}}%
      \put(0.69438854,0.15942759){\makebox(0,0)[lb]{\smash{0}}}%
      \put(0.69438854,0.11620775){\makebox(0,0)[lb]{\smash{0}}}%
      \put(0.69438854,0.07298791){\makebox(0,0)[lb]{\smash{0}}}%
      \put(0.73760834,0.80772522){\makebox(0,0)[lb]{\smash{0}}}%
      \put(0.73760834,0.76450542){\makebox(0,0)[lb]{\smash{0}}}%
      \put(0.73760834,0.72128554){\makebox(0,0)[lb]{\smash{0}}}%
      \put(0.73760834,0.67806574){\makebox(0,0)[lb]{\smash{0}}}%
      \put(0.73760834,0.63484587){\makebox(0,0)[lb]{\smash{0}}}%
      \put(0.73760834,0.59162603){\makebox(0,0)[lb]{\smash{0}}}%
      \put(0.72954108,0.5484062){\makebox(0,0)[lb]{\smash{0.5}}}%
      \put(0.73760834,0.50518636){\makebox(0,0)[lb]{\smash{0}}}%
      \put(0.72954108,0.46196649){\makebox(0,0)[lb]{\smash{3.9}}}%
      \put(0.72954108,0.41874665){\makebox(0,0)[lb]{\smash{0.5}}}%
      \put(0.73760834,0.37552681){\makebox(0,0)[lb]{\smash{0}}}%
      \put(0.73760834,0.33230697){\makebox(0,0)[lb]{\smash{0}}}%
      \put(0.73760834,0.28908712){\makebox(0,0)[lb]{\smash{0}}}%
      \put(0.73760834,0.24586728){\makebox(0,0)[lb]{\smash{0}}}%
      \put(0.73760834,0.20264745){\makebox(0,0)[lb]{\smash{0}}}%
      \put(0.73760834,0.15942759){\makebox(0,0)[lb]{\smash{0}}}%
      \put(0.72435501,0.11620775){\makebox(0,0)[lb]{\color{white}\smash{95.1}}}%
      \put(0.73760834,0.07298791){\makebox(0,0)[lb]{\smash{0}}}%
      \put(0.80023486,0.02531474){\makebox(0,0)[lb]{\smash{1}}}%
      \put(0.84345473,0.02531474){\makebox(0,0)[lb]{\smash{2}}}%
      \put(0.80108752,0.80291155){\makebox(0,0)[lb]{\smash{0}}}%
      \put(0.79302026,0.75969174){\makebox(0,0)[lb]{\smash{0.3}}}%
      \put(0.78783412,0.71647187){\makebox(0,0)[lb]{\smash{11.7}}}%
      \put(0.80108752,0.67325207){\makebox(0,0)[lb]{\smash{0}}}%
      \put(0.80108752,0.6300322){\makebox(0,0)[lb]{\smash{0}}}%
      \put(0.79302026,0.58681236){\makebox(0,0)[lb]{\smash{0.2}}}%
      \put(0.79302026,0.54359252){\makebox(0,0)[lb]{\smash{1.4}}}%
      \put(0.78783412,0.50037269){\makebox(0,0)[lb]{\smash{11.1}}}%
      \put(0.80108752,0.45715282){\makebox(0,0)[lb]{\smash{0}}}%
      \put(0.79302026,0.41393298){\makebox(0,0)[lb]{\smash{0.4}}}%
      \put(0.78783412,0.37071314){\makebox(0,0)[lb]{\smash{40.7}}}%
      \put(0.80108752,0.3274933){\makebox(0,0)[lb]{\smash{0}}}%
      \put(0.80108752,0.28427345){\makebox(0,0)[lb]{\smash{0}}}%
      \put(0.80108752,0.24105361){\makebox(0,0)[lb]{\smash{0}}}%
      \put(0.78783412,0.19783378){\makebox(0,0)[lb]{\smash{22.2}}}%
      \put(0.78783412,0.15461392){\makebox(0,0)[lb]{\smash{12.0}}}%
      \put(0.80108752,0.11139408){\makebox(0,0)[lb]{\smash{0}}}%
      \put(0.79302026,0.06817424){\makebox(0,0)[lb]{\smash{0.1}}}%
      \put(0.83624013,0.80291155){\makebox(0,0)[lb]{\smash{0.6}}}%
      \put(0.83624013,0.75969174){\makebox(0,0)[lb]{\smash{1.8}}}%
      \put(0.83624013,0.71647187){\makebox(0,0)[lb]{\smash{0.1}}}%
      \put(0.83624013,0.67325207){\makebox(0,0)[lb]{\smash{2.0}}}%
      \put(0.83624013,0.6300322){\makebox(0,0)[lb]{\smash{1.0}}}%
      \put(0.83624013,0.58681236){\makebox(0,0)[lb]{\smash{7.7}}}%
      \put(0.83624013,0.54359252){\makebox(0,0)[lb]{\smash{0.1}}}%
      \put(0.83624013,0.50037269){\makebox(0,0)[lb]{\smash{0.3}}}%
      \put(0.83624013,0.45715282){\makebox(0,0)[lb]{\smash{2.1}}}%
      \put(0.83624013,0.41393298){\makebox(0,0)[lb]{\smash{0.1}}}%
      \put(0.83624013,0.37071314){\makebox(0,0)[lb]{\smash{0.1}}}%
      \put(0.83624013,0.3274933){\makebox(0,0)[lb]{\smash{0.1}}}%
      \put(0.84430739,0.28427345){\makebox(0,0)[lb]{\smash{0}}}%
      \put(0.83105399,0.24105361){\makebox(0,0)[lb]{\smash{49.2}}}%
      \put(0.83624013,0.19783378){\makebox(0,0)[lb]{\smash{0.1}}}%
      \put(0.84430739,0.15461392){\makebox(0,0)[lb]{\smash{0}}}%
      \put(0.83105399,0.11139408){\makebox(0,0)[lb]{\smash{16.3}}}%
      \put(0.83105399,0.06817424){\makebox(0,0)[lb]{\smash{18.6}}}%
      \put(0.9692484,0.04533485){\makebox(0,0)[lb]{\smash{0}}}%
      \put(0.98132586,0.04533485){\makebox(0,0)[lb]{\smash{.}}}%
      \put(0.98803555,0.04533485){\makebox(0,0)[lb]{\smash{0}}}%
      \put(0.9692484,0.12313057){\makebox(0,0)[lb]{\smash{0}}}%
      \put(0.98132586,0.12313057){\makebox(0,0)[lb]{\smash{.}}}%
      \put(0.98803555,0.12313057){\makebox(0,0)[lb]{\smash{1}}}%
      \put(0.9692484,0.20092628){\makebox(0,0)[lb]{\smash{0}}}%
      \put(0.98132586,0.20092628){\makebox(0,0)[lb]{\smash{.}}}%
      \put(0.98803555,0.20092628){\makebox(0,0)[lb]{\smash{2}}}%
      \put(0.9692484,0.278722){\makebox(0,0)[lb]{\smash{0}}}%
      \put(0.98132586,0.278722){\makebox(0,0)[lb]{\smash{.}}}%
      \put(0.98803555,0.278722){\makebox(0,0)[lb]{\smash{3}}}%
      \put(0.9692484,0.35651771){\makebox(0,0)[lb]{\smash{0}}}%
      \put(0.98132586,0.35651771){\makebox(0,0)[lb]{\smash{.}}}%
      \put(0.98803555,0.35651771){\makebox(0,0)[lb]{\smash{4}}}%
      \put(0.9692484,0.43431343){\makebox(0,0)[lb]{\smash{0}}}%
      \put(0.98132586,0.43431343){\makebox(0,0)[lb]{\smash{.}}}%
      \put(0.98803555,0.43431343){\makebox(0,0)[lb]{\smash{5}}}%
      \put(0.9692484,0.51210914){\makebox(0,0)[lb]{\smash{0}}}%
      \put(0.98132586,0.51210914){\makebox(0,0)[lb]{\smash{.}}}%
      \put(0.98803555,0.51210914){\makebox(0,0)[lb]{\smash{6}}}%
      \put(0.9692484,0.58990489){\makebox(0,0)[lb]{\smash{0}}}%
      \put(0.98132586,0.58990489){\makebox(0,0)[lb]{\smash{.}}}%
      \put(0.98803555,0.58990489){\makebox(0,0)[lb]{\smash{7}}}%
      \put(0.9692484,0.6677006){\makebox(0,0)[lb]{\smash{0}}}%
      \put(0.98132586,0.6677006){\makebox(0,0)[lb]{\smash{.}}}%
      \put(0.98803555,0.6677006){\makebox(0,0)[lb]{\smash{8}}}%
      \put(0.9692484,0.74549632){\makebox(0,0)[lb]{\smash{0}}}%
      \put(0.98132586,0.74549632){\makebox(0,0)[lb]{\smash{.}}}%
      \put(0.98803555,0.74549632){\makebox(0,0)[lb]{\smash{9}}}%
      \put(0.9692484,0.82329203){\makebox(0,0)[lb]{\smash{1}}}%
      \put(0.98132586,0.82329203){\makebox(0,0)[lb]{\smash{.}}}%
      \put(0.98803555,0.82329203){\makebox(0,0)[lb]{\smash{0}}}}%
      {\fontsize{8pt}{1em} \selectfont%
      \put(0.00983972,0.37872909){\color[rgb]{0,0,0}\rotatebox{90}{\makebox(0,0)[lb]{\smash{Primitive actions}}}}%
      \put(0.1,0.85072122){\color[rgb]{0,0,0}\makebox(0,0)[lb]{\smash{8 options, no baseline}}}%
      \put(0.45,0.85072122){\color[rgb]{0,0,0}\makebox(0,0)[lb]{\smash{8 options, baseline}}}%
      \put(0.78,0.85072122){\color[rgb]{0,0,0}\makebox(0,0)[lb]{\smash{2 options, baseline}}}%
      }%
    \end{picture}%
  \endgroup%
  \caption{Seaquest: Using a baseline in the gradient estimators improves the distribution
  over actions in the intra-option policies, making them less deterministic.
  Each column represents one of the options learned in Seaquest.
  The vertical axis spans the $18$ primitive actions of ALE. The empirical action
  frequencies are coded by intensity.}
  \label{fig:seaquest_baseline}
\end{figure}
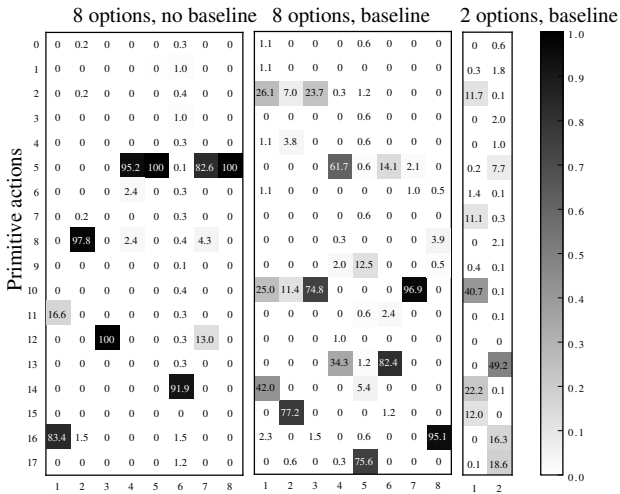

Finally, the baseline $Q_\Omega$ was added to the intra-option policy gradient estimator
to reduce its variance.
This change provided substantial improvements \cite{Harb2016} in the quality of the
intra-option policy distributions and the overall agent performance as explained
in Figure \ref{fig:seaquest_baseline}.

\begin{figure*}[tb]
\centering
\def\svgwidth{0.9\textwidth}
\begingroup  \makeatletter  \providecommand\color[2][]{    \errmessage{(Inkscape) Color is used for the text in Inkscape, but the package 'color.sty' is not loaded}    \renewcommand\color[2][]{}  }  \providecommand\transparent[1]{    \errmessage{(Inkscape) Transparency is used (non-zero) for the text in Inkscape, but the package 'transparent.sty' is not loaded}    \renewcommand\transparent[1]{}  }  \providecommand\rotatebox[2]{#2}  \ifx\svgwidth\undefined    \setlength{\unitlength}{505.88999023bp}    \ifx\svgscale\undefined      \relax    \else      \setlength{\unitlength}{\unitlength * \real{\svgscale}}    \fi  \else    \setlength{\unitlength}{\svgwidth}  \fi  \global\let\svgwidth\undefined  \global\let\svgscale\undefined  \makeatother  \begin{picture}(1,0.20391707)    \put(0,0){\includegraphics[width=\unitlength]{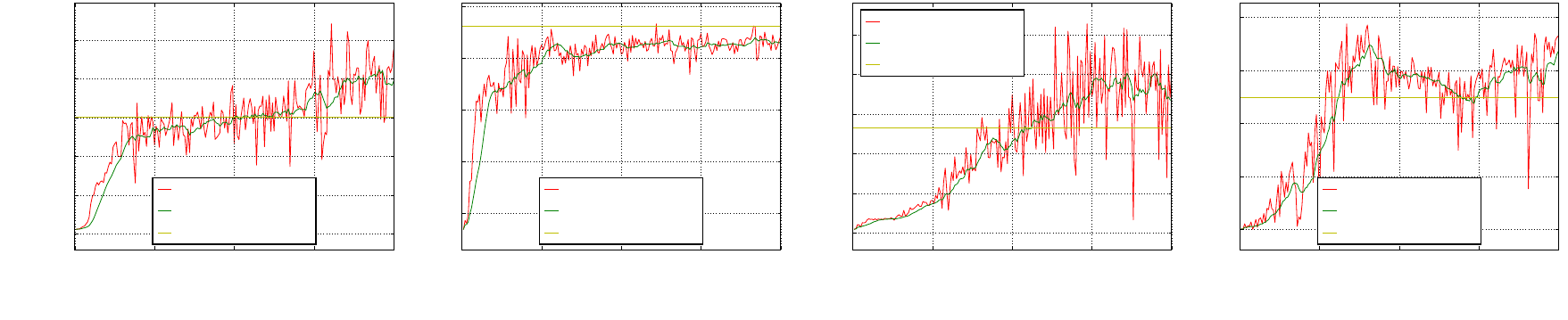}}    {\fontsize{9pt}{1em} \selectfont    \put(0.10964097,0.001){\color[rgb]{0,0,0}\makebox(0,0)[lb]{\smash{(a) Asterix}}}    \put(0.35356528,0.001){\color[rgb]{0,0,0}\makebox(0,0)[lb]{\smash{(b) Ms. Pacman}}}    \put(0.5917876,0.001){\color[rgb]{0,0,0}\makebox(0,0)[lb]{\smash{(c) Seaquest}}}    \put(0.85345847,0.001){\color[rgb]{0,0,0}\makebox(0,0)[lb]{\smash{(d) Zaxxon}}}    }    \put(0.11451793,0.07980189){{\fontsize{6pt}{1em} \selectfont Testing}}    \put(0.11450534,0.06607529){{\fontsize{6pt}{1em} \selectfont Moving avg.10}}    \put(0.11453047,0.05217269){{\fontsize{6pt}{1em} \selectfont DQN}}    \put(0.36144164,0.07980189){{\fontsize{6pt}{1em} \selectfont Testing}}    \put(0.36142905,0.06607529){{\fontsize{6pt}{1em} \selectfont Moving avg.10}}    \put(0.36145419,0.05217269){{\fontsize{6pt}{1em} \selectfont DQN}}    \put(0.56621405,0.18725355){{\fontsize{6pt}{1em} \selectfont Testing}}    \put(0.56620146,0.17352694){{\fontsize{6pt}{1em} \selectfont Moving avg.10}}    \put(0.56622658,0.15962434){{\fontsize{6pt}{1em} \selectfont DQN}}    \put(0.85773333,0.07980189){{\fontsize{6pt}{1em} \selectfont Testing}}    \put(0.85772074,0.06607529){{\fontsize{6pt}{1em} \selectfont Moving avg.10}}    \put(0.85774586,0.05217269){{\fontsize{6pt}{1em} \selectfont DQN}}    {\fontsize{6pt}{1em} \selectfont    \put(0.0101261,0.09045289){\color[rgb]{0,0,0}\rotatebox{90}{\makebox(0,0)[lb]{\smash{Avg. Score}}}}    \put(0.13194979,0.01967973){\color[rgb]{0,0,0}\makebox(0,0)[lb]{\smash{Epoch}}}    \put(0.37887354,0.01967973){\color[rgb]{0,0,0}\makebox(0,0)[lb]{\smash{Epoch}}}    \put(0.62824148,0.01967973){\color[rgb]{0,0,0}\makebox(0,0)[lb]{\smash{Epoch}}}    \put(0.8751652,0.01967973){\color[rgb]{0,0,0}\makebox(0,0)[lb]{\smash{Epoch}}}    \put(0.08910699,0.03295184){\color[rgb]{0,0,0}\makebox(0,0)[lb]{\smash{50}}}    \put(0.13630881,0.03295184){\color[rgb]{0,0,0}\makebox(0,0)[lb]{\smash{100 }}}    \put(0.18740138,0.03295184){\color[rgb]{0,0,0}\makebox(0,0)[lb]{\smash{150}}}    \put(0.23843071,0.03295184){\color[rgb]{0,0,0}\makebox(0,0)[lb]{\smash{200}}}    \put(0.04153471,0.03295184){\color[rgb]{0,0,0}\makebox(0,0)[lb]{\smash{0}}}    \put(0.33603069,0.03295184){\color[rgb]{0,0,0}\makebox(0,0)[lb]{\smash{50}}}    \put(0.3832325,0.03295184){\color[rgb]{0,0,0}\makebox(0,0)[lb]{\smash{100 }}}    \put(0.43432508,0.03295184){\color[rgb]{0,0,0}\makebox(0,0)[lb]{\smash{150}}}    \put(0.48535441,0.03295184){\color[rgb]{0,0,0}\makebox(0,0)[lb]{\smash{200}}}    \put(0.28845842,0.03295184){\color[rgb]{0,0,0}\makebox(0,0)[lb]{\smash{0}}}    \put(0.58539864,0.03295184){\color[rgb]{0,0,0}\makebox(0,0)[lb]{\smash{50}}}    \put(0.63260044,0.03295184){\color[rgb]{0,0,0}\makebox(0,0)[lb]{\smash{100 }}}    \put(0.68369303,0.03295184){\color[rgb]{0,0,0}\makebox(0,0)[lb]{\smash{150}}}    \put(0.73472236,0.03295184){\color[rgb]{0,0,0}\makebox(0,0)[lb]{\smash{200}}}    \put(0.53782636,0.03295184){\color[rgb]{0,0,0}\makebox(0,0)[lb]{\smash{0}}}    \put(0.83232235,0.03295184){\color[rgb]{0,0,0}\makebox(0,0)[lb]{\smash{50}}}    \put(0.87952416,0.03295184){\color[rgb]{0,0,0}\makebox(0,0)[lb]{\smash{100 }}}    \put(0.93061674,0.03295184){\color[rgb]{0,0,0}\makebox(0,0)[lb]{\smash{150}}}    \put(0.98164607,0.03295184){\color[rgb]{0,0,0}\makebox(0,0)[lb]{\smash{200}}}    \put(0.78475007,0.03295184){\color[rgb]{0,0,0}\makebox(0,0)[lb]{\smash{0}}}    \put(0.0379396,0.05061234){\color[rgb]{0,0,0}\makebox(0,0)[lb]{\smash{0}}}    \put(0.01908281,0.07530572){\color[rgb]{0,0,0}\makebox(0,0)[lb]{\smash{2000}}}    \put(0.01908281,0.09992981){\color[rgb]{0,0,0}\makebox(0,0)[lb]{\smash{4000}}}    \put(0.01908281,0.12469237){\color[rgb]{0,0,0}\makebox(0,0)[lb]{\smash{6000}}}    \put(0.01908281,0.14938564){\color[rgb]{0,0,0}\makebox(0,0)[lb]{\smash{8000}}}    \put(0.01279722,0.17407891){\color[rgb]{0,0,0}\makebox(0,0)[lb]{\smash{10000}}}    \put(0.27205433,0.0635936){\color[rgb]{0,0,0}\makebox(0,0)[lb]{\smash{500}}}    \put(0.26576874,0.09658134){\color[rgb]{0,0,0}\makebox(0,0)[lb]{\smash{1000}}}    \put(0.26576874,0.12956907){\color[rgb]{0,0,0}\makebox(0,0)[lb]{\smash{1500}}}    \put(0.26576874,0.16255692){\color[rgb]{0,0,0}\makebox(0,0)[lb]{\smash{2000}}}    \put(0.26576874,0.19554466){\color[rgb]{0,0,0}\makebox(0,0)[lb]{\smash{2500}}}    \put(0.53495811,0.05061234){\color[rgb]{0,0,0}\makebox(0,0)[lb]{\smash{0}}}    \put(0.51610132,0.07530572){\color[rgb]{0,0,0}\makebox(0,0)[lb]{\smash{2000}}}    \put(0.51610132,0.09992981){\color[rgb]{0,0,0}\makebox(0,0)[lb]{\smash{4000}}}    \put(0.51610132,0.12469237){\color[rgb]{0,0,0}\makebox(0,0)[lb]{\smash{6000}}}    \put(0.51610132,0.14938564){\color[rgb]{0,0,0}\makebox(0,0)[lb]{\smash{8000}}}    \put(0.50981572,0.17407891){\color[rgb]{0,0,0}\makebox(0,0)[lb]{\smash{10000}}}    \put(0.78032843,0.05337097){\color[rgb]{0,0,0}\makebox(0,0)[lb]{\smash{0}}}    \put(0.76147163,0.0872382){\color[rgb]{0,0,0}\makebox(0,0)[lb]{\smash{2000}}}    \put(0.76147163,0.12103614){\color[rgb]{0,0,0}\makebox(0,0)[lb]{\smash{4000}}}    \put(0.76147163,0.15497243){\color[rgb]{0,0,0}\makebox(0,0)[lb]{\smash{6000}}}    \put(0.76147163,0.18883955){\color[rgb]{0,0,0}\makebox(0,0)[lb]{\smash{8000}}}}  \end{picture}\endgroup   \caption{Learning curves in the Arcade Learning Environment. The same set of
  parameters was used across all four games: $8$ options, $0.01$ termination
  regularization, $0.01$ entropy regularization, and a baseline for the
  intra-option policy gradients.}
  \label{fig:many_games_performance_OC}
\end{figure*}
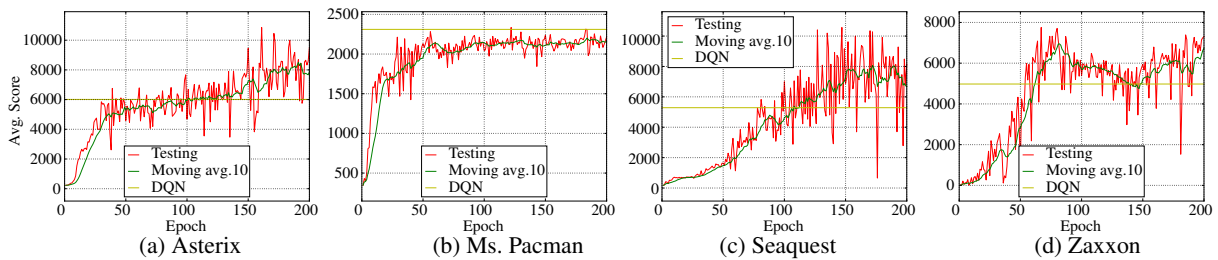

We evaluated option-critic
in \textit{Asterisk}, \textit{Ms. Pacman}, \textit{Seaquest} and \textit{Zaxxon}.
For comparison, we allowed the system to learn for the same number of
episodes as \cite{Mnih2013} and fixed the parameters to the same values in all
four domains.  Despite having more parameters to learn, option-critic was capable
of learning options that would achieve the goal in all games, from the ground up, within 200 episodes (Figure \ref{fig:many_games_performance_OC}).
In Asterisk, Seaquest and Zaxxon, option-critic surpassed the performance of
the original DQN architecture based on primitive actions. The eight options learned in each
game are learned fully end-to-end, in tandem with the feature representation,
with no prior specification of a subgoal or pseudo-reward structure.

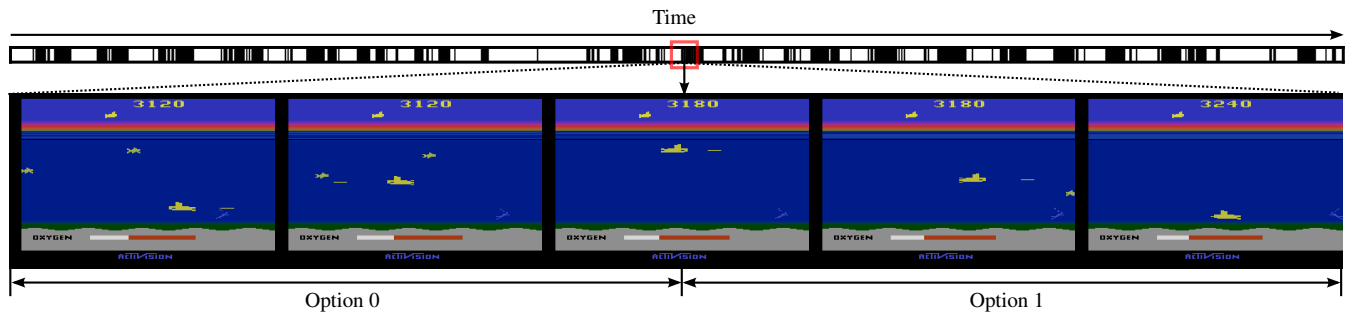
\begin{figure*}[tb]
  \def\svgwidth{\textwidth}
  \begingroup  \makeatletter  \providecommand\color[2][]{    \errmessage{(Inkscape) Color is used for the text in Inkscape, but the package 'color.sty' is not loaded}    \renewcommand\color[2][]{}  }  \providecommand\transparent[1]{    \errmessage{(Inkscape) Transparency is used (non-zero) for the text in Inkscape, but the package 'transparent.sty' is not loaded}    \renewcommand\transparent[1]{}  }  \providecommand\rotatebox[2]{#2}  \ifx\svgwidth\undefined    \setlength{\unitlength}{505.01323242bp}    \ifx\svgscale\undefined      \relax    \else      \setlength{\unitlength}{\unitlength * \real{\svgscale}}    \fi  \else    \setlength{\unitlength}{\svgwidth}  \fi  \global\let\svgwidth\undefined  \global\let\svgscale\undefined  \makeatother  {\fontsize{8pt}{0em} \selectfont  \begin{picture}(1,0.22593052)    \put(0,0){\includegraphics[width=\unitlength]{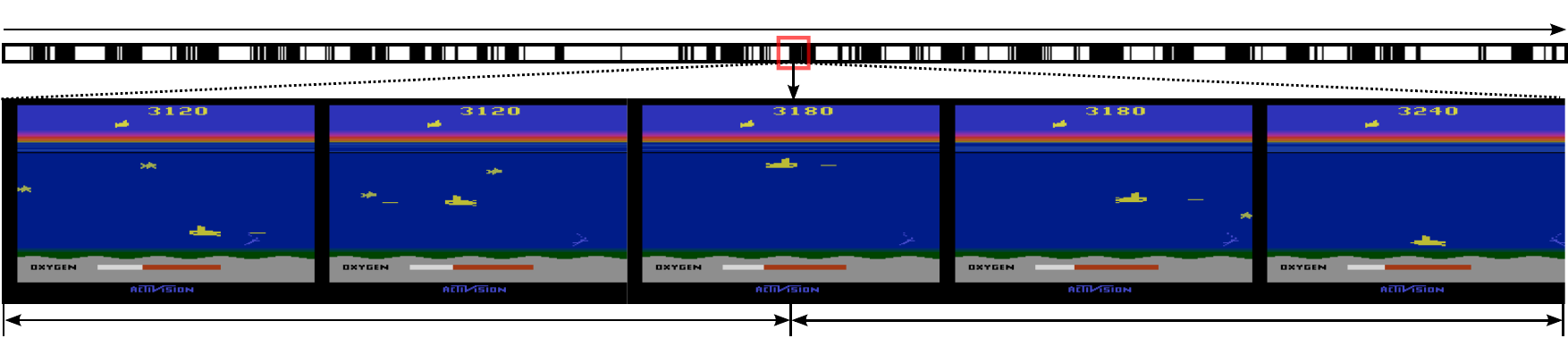}}    \put(0.22266124,0.00310882){\color[rgb]{0,0,0}\makebox(0,0)[lb]{\smash{Option 0}}}    \put(0.71967768,0.00310882){\color[rgb]{0,0,0}\makebox(0,0)[lb]{\smash{Option 1}}}    \put(0.48232898,0.21508169){\color[rgb]{0,0,0}\makebox(0,0)[lb]{\smash{Time}}}  \end{picture}}\endgroup   \caption{Up/down specialization in the solution found by option-critic when learning with 2 options in Seaquest.
  The top bar shows a trajectory in the game, with ``white'' representing
  a segment during which option 1 was active and ``black'' for option 2.}
  \label{fig:seaquest_2opt}
\end{figure*}

The solution found by option-critic was easy to interpret in
the game of Seaquest when learning with only two options. We found that each
option specialized in a behavior sequence which would include either the up
or the down button.
Figure \ref{fig:seaquest_2opt} shows a typical transition
from one option to the other, first going upward with option $0$ then switching
to option $1$ downward. Options with a similar structure were also found in this game by
\cite{Krishnamurthy2016} using an option discovery algorithm based on graph partitioning.

 \section{Related Work}
As option discovery has received a lot of attention recently, we now discuss in
more detail the place of our approach with respect to others.
\cite{Comanici2010} used a gradient-based approach for improving only the termination
function of semi-Markov options; termination was modeled by a logistic
distribution over a cumulative measure of the features observed since initiation.
\cite{levy2011} also built on policy gradient methods by constructing explicitly the augmented
state space and treating stopping events as additional control actions.
In contrast, we do not need to construct this (very large) space directly.
\cite{Silver2012} dynamically chained options into longer temporal sequences by
relying on compositionality properties. Earlier work on linear
options~\cite{Sorg2010} also used compositionality to plan using linear
expectation models for options.
Our approach also relies on the Bellman equations and compositionality, but in conjunction with policy gradient methods.

Several very recent papers also attempt to formulate option discovery as an
optimization problem with solutions that are compatible with function approximation.
\cite{Daniel2016} learn return-optimizing options by treating the termination
functions as hidden variables, and using EM to learn them. \cite{Vezhnevets2016}
consider the problem of learning options that have open-loop intra-option policies,
also called \textit{macro-actions}.
As in classical planning, action sequences that are more frequent are cached.
A mapping from states to action sequences
is learned along with a \textit{commitment module}, which triggers
re-planning when necessary. In contrast, we use closed-loop policies throughout,
which are reactive to state information and can provide better solutions.
\cite{Mankowitz2016} propose a gradient-based option learning algorithm,
assuming a particular structure for the initiation sets and termination functions.
Under this framework, exactly one option is active in any partition of the state space.
\cite{Kulkarni2016} use the DQN framework to implement a gradient-based option
learner, which uses intrinsic rewards to learn the internal policies of options,
and extrinsic rewards to learn the policy over options. As opposed to our framework,
descriptions of the subgoals are given as inputs to the option learners.
Option-critic is conceptually general and does not require intrinsic motivation
for learning the options.

\section{Discussion}\label{sect:discussion}

We developed a general gradient-based approach for learning simultaneously the
intra-option policies and termination functions, as well as the policy over options,
in order to optimize a performance objective for the task at hand.
Our ALE experiments demonstrate successful
end-to-end learning of options in the presence of nonlinear function
approximation. As noted,  our approach only requires specifying
the number of options.
However, if one wanted to use additional
pseudo-rewards, the option-critic framework would easily accommodate it.
In this case, the internal policies and termination function gradients would simply need to be
taken with respect to the pseudo-rewards instead of the task reward.
A simple instance of this idea, which we used in some of the experiments, is to
use additional rewards to encourage options that are indeed temporally
extended by adding a penalty whenever a switching event
occurs. Our approach can work seamlessly with any other heuristic for biasing
the set of options towards some desirable property (e.g. compositionality or sparsity),
as long as it can be expressed as an additive reward structure.
However, as seen in the results, such biasing is not necessary
to produce good results.

The option-critic architecture relies on the policy gradient theorem, and as discussed in
 \cite{thomas2014bias}, the gradient estimators can be biased
in the discounted case. By introducing factors of the form
$\gamma^t  \prod_{i=1}^t(1-\beta_i)$ in our updates \cite[eq (3)]{thomas2014bias},
it would be possible to obtain unbiased estimates. However, we do not recommend
this approach since the sample complexity of the unbiased estimators is generally
too high and the biased estimators performed well in our experiments.

Perhaps the biggest remaining limitation of our work is the assumption that all
options apply everywhere.  In the case of function approximation, a natural
extension to initiation sets is to use a classifier over features, or some other
form of function approximation.  As a result, determining which options are
allowed may have similar cost to evaluating a policy over options
(unlike in the tabular setting, where options with sparse initiation sets lead to faster decisions).
This is akin to eligibility traces, which are more expensive than using no
trace in the tabular case, but have the same complexity with function approximation.
If initiation sets are to be learned, the main constraint that needs to be added
is that the options and the policy over them lead to an ergodic chain in the augmented
state-option space. This can be expressed as a flow condition that links
initiation sets with terminations. The precise description of this condition,
as well as sparsity regularization for initiation sets, is left for future work.

\section{Acknowledgements}
The authors gratefully acknowledge financial support for this work by the National
Science and Engineering Research Council of Canada (NSERC) and the Fonds de
recherche du Quebec - Nature et Technologies (FRQNT).

\section{Appendix}

\subsection*{Augmented Process}
If $\omega_t$ has been
initiated or is executing at time $t$,
then the discounted probability of transitioning to $(s_{t+1}, \omega_{t+1})$ is:
\begin{align*}
&\prob^{(1)}_\gamma\left(s_{t+1}, \omega_{t+1} |\, s_t, \omega_t \right) = \sum_{a} \pi_{\omega_t}\left(a |\, s_t \right)\gamma \prob(s_{t+1} |\, s_t, a)\big( \\
& (1 - \beta_{\omega_t}(s_{t+1}))\indicator_{\omega_{t}=\omega_{t+1}} \
+ \beta_{\omega_{t}}(s_{t+1})\pi_\Omega\left(\omega_{t+1} \given s_{t+1}\right)\big) \enspace .
\end{align*}
When conditioning the process from $(s_t, \omega_{t-1})$, the discounted probability of transitioning to
$s_{t+1}, \omega_{t}$ is:
\begin{align*}
&\prob^{(1)}_\gamma\left(s_{t+1}, \omega_{t} \given s_t, \omega_{t-1} \right) = \big( (1 - \beta_{\omega_{t-1}}(s_{t}))\indicator_{\omega_{t}=\omega_{t-1}} + \\
&\beta_{\omega_{t-1}}(s_{t})\pi_\Omega\left(\omega_{t} \given s_{t}\right)\big)
 \sum_{a} \pi_{\omega_t}\left(a \given s_{t}\right) \gamma \prob\left(s_{t+1} \given s_t, a\right) \enspace .
\end{align*}
More generally, the $k$-steps discounted probabilities can be expressed
recursively as follows:
\begin{align*}
&\prob^{(k)}_\gamma\left(s_{t+k}, \omega_{t+k}\given s_t, \omega_{t}\right) = \sum_{s_{t+1}} \sum_{\option_{t+1}} \big( \\
& \prob_\gamma^{(1)}\left( s_{t+1}, \omega_{t+1} \given s_{t}, \omega_{t} \right)\prob_\gamma^{(k-1)}\left( s_{t+k}, \omega_{t+k} \given s_{t+1}, \omega_{t+1} \right) \big) \, ,\\
&\prob^{(k)}_\gamma\left(s_{t+k}, \omega_{t+k-1}\given s_t, \omega_{t-1}\right) = \sum_{s_{t+1}} \sum_{\option_{t}} \big( \\
&\prob_\gamma^{(1)}\left( s_{t+1}, \omega_t \given s_{t}, \omega_{t-1} \right) \prob_\gamma^{(k-1)}\left( s_{t+k}, \omega_{t+k-1} \given s_{t+1}, \omega_{t} \right) \big) \, .
\end{align*}

\subsection*{Proof of the Intra-Option Policy Gradient Theorem}
Taking the gradient of the option-value function:
\begin{align}
&\deriv[Q_\Omega(s, \omega)]{\theta} = \deriv[]{\theta} \sum_a \pi_{\omega,\theta}\left( a \given s\right) Q_U(s,\omega, a)\notag\\
&= \sum_a \Bigg(\deriv[\pi_{\omega,\theta}\left( a | s \right)]{\theta}  Q_U(s,\omega, a) + \notag \\
&\hspace{4em}\pi_{\omega,\theta}\left(a | s\right)\deriv[Q_U(s,\omega, a)]{\theta} \Bigg) \notag \\
&= \sum_a \Bigg(\deriv[\pi_{\omega,\theta}\left( a \given s \right)]{\theta} Q_U(s,\omega, a) + \notag \\
&\hspace{4em} \pi_{\omega,\theta}\left(a \given s\right)\sum_{s^\prime} \gamma \prob\left( s^\prime \given s, a\right)\deriv[U(\omega, s')]{\theta} \Bigg) \enspace , \label{eq:pg}\\
&\deriv[U(\omega, s')]{\theta} = \notag\\
&(1 - \beta_{\omega, \vartheta}(s')) \deriv[Q_\Omega(s', \omega)]{\theta} + \beta_{\omega, \vartheta}(s')\deriv[V_\Omega(s')]{\theta} \notag \\
&= (1 - \beta_{\omega, \vartheta}(s'))\deriv[Q_\Omega(s', \omega)]{\theta} + \notag \\
&\hspace{4em}\beta_{\omega, \vartheta}(s') \sum_{\omega^\prime} \pi_\Omega \left( \omega^\prime \given s' \right)\deriv[ Q_\Omega(s', \omega^\prime)]{\theta} \notag \\
&= \sum_{\omega'}\big((1 -\beta_{\omega, \vartheta}(s'))\indicator_{\omega'=\omega} + \notag \\
&\hspace{4em}\beta_{\omega, \vartheta}(s') \pi_\Omega\left(\omega' \given s'\right)\big)\
  \deriv[Q_\Omega(s', \omega')]{\theta} \enspace .  \label{eq:upg}
\end{align}
where \eqref{eq:upg} follows from the assumption that $\theta$ only appears in the intra-option policies.
Substituting  \eqref{eq:upg} into \eqref{eq:pg} yields a recursion which, using
the previous remarks about augmented process
can be transformed into:
\begin{align*}
&\deriv[Q_\Omega(s, \omega)]{\theta} = \sum_a \deriv[\pi_{\omega,\theta}\left( a \given s \right)]{\theta}  Q_U(s,\omega, a) + \\
&\sum_a \pi_{\omega, \theta}\left( a \given s\right) \sum_{s'} \gamma \prob\left(s' \given s, a\right) \sum_{\omega'}\Big(\beta_{\omega, \vartheta}(s') \pi_\Omega\left(\omega' \given s'\right) \notag\\
&\hspace{4em}+ (1 -\beta_{\omega, \vartheta}(s'))\indicator_{\omega'=\omega}  \Big) \deriv[Q_\Omega(s', \omega')]{\theta} \notag \\
&= \sum_a \deriv[\pi_{\omega,\theta}\left( a \given s \right)]{\theta}  Q_U(s,\omega, a) + \notag \\
&\hspace{4em}\sum_{s'} \sum_{\omega'} \prob_\gamma^{(1)}\left( s', \omega' \given s, \omega\right) \deriv[Q_\Omega(s', \omega')]{\theta}\\
&= \sum_{k=0}^\infty \sum_{s',\omega'} \prob_\gamma^{(k)}\left( s', \omega' | s, \omega\right) \sum_a \deriv[\pi_{\omega',\theta}\left( a | s' \right)]{\theta}  Q_U(s',\omega', a)  .
\end{align*}
The gradient of the expected discounted return with respect to $\theta$ is then:
\begin{align*}
&\deriv[Q_\Omega(s_0, \omega_0)]{\theta} = \\
&\sum_{s,\omega} \sum_{k=0}^\infty \prob_\gamma^{(k)}\left( s, \omega \given s_0, \omega_0\right)\sum_a \deriv[\pi_{\omega,\theta}\left( a \given s \right)]{\theta}Q_U(s,\omega, a) \\
&= \sum_{s, \omega} \mu_\Omega(s, \omega | s_0, \omega_0) \sum_a \deriv[\pi_{\omega,\theta}\left( a \given s \right)]{\theta} Q_U(s,\omega, a) \enspace .
\end{align*}
\subsection*{Proof of the Termination Gradient Theorem}
The expected sum of discounted rewards starting from $(s_1, \omega_0)$ is given by:
\begin{align*}
U(\omega_0, s_1) = \expectation\left[ \sum_{t=1}^\infty \gamma^{t-1} r_{t}  \given s_1, \omega_0 \right] \enspace .
\end{align*}
We start by expanding $U$ as follows:
\begin{align}
&U( \option, s') = (1 - \beta_{\option, \termparams}(s'))Q_\Omega(s', \omega) + \beta_{\option, \termparams}(s')V_\Omega(s') \notag\\
&=(1 - \beta_{\option, \termparams}(s')) \sum_{a} \pi_{\option,\theta}\left(a \given s'\right) \Big(\notag \\
&\hspace{4em}r(s', a) + \sum_{s''} \gamma \prob\left(s'' \given s', a \right)U(\option, s'')\Big) \notag \\
&+ \;\beta_{\option, \termparams}(s') \sum_{\option'}\pi_{\Omega}\left( \option' \given s'\right)\sum_a \pi_{\option',\theta}\left(a \given s' \right)\Big( \notag \\
&\hspace{4em} r(s', a) + \sum_{s''} \gamma  \prob\left(s'' \given s', a\right) U(\option', s'')\Big)\notag \enspace .
\end{align}
The gradient of $U$ is then:
\begin{align*}
&\deriv[U(\option, s')]{\termparams} = \deriv[\beta_{\option, \termparams}(s')]{\termparams}\underbrace{\left(V_\options(s') - Q_\options(s', \option)\right)}_{-A_\options(s', \omega)} +\\
&(1 - \beta_{\option, \termparams}(s'))\sum_a \pi_{\option,\theta}\left(a | s'\right)\sum_{s''} \gamma \prob\left(s'' | s', a \right) \deriv[U(\option, s'')]{\termparams} .\notag\\
\end{align*}
Using the structure of the augmented process:
\begin{align*}
&\deriv[U(\option, s')]{\termparams} = -\deriv[\beta_{\option, \termparams}(s')]{\termparams} A_\options(s', \option) + \\
&\hspace{4em}\sum_{\option'} \sum_{s''} \prob_\gamma^{(1)}\left(s'', \option' \given s', \omega\right) \deriv[U(\option', s'')]{\termparams} \\
&=- \sum_{\option', s''} \sum_{k=0}^\infty \prob_\gamma^{(k)}\left(s'', \option' \given s',\omega \right) \deriv[\beta_{\option', \termparams}(s'')]{\termparams} A_\options(s'', \option') \enspace .
\end{align*}
 We finally obtain:
\begin{align*}
&\deriv[U(\omega_0, s_1)]{\vartheta} = \\
&-\sum_{\option, s'} \sum_{k=0}^\infty  \prob_\gamma^{(k)}\left(s', \option \given s_1, \omega_0 \right) \deriv[\beta_{\option, \termparams}(s')]{\termparams} A_\options(s', \option) \\
&=-\sum_{\option, s'} \mu_\Omega(s', \omega | s_1, \omega_0) \deriv[\beta_{\option, \termparams}(s')]{\termparams} A_\options(s', \option) \enspace.
\end{align*}

 \bibliographystyle{aaai}
 \bibliography{references}

\end{document}